\documentclass[runningheads]{llncs}
\usepackage[T1]{fontenc}
\usepackage{graphicx}

\usepackage{amssymb}
\usepackage{amsmath,amsfonts}
\usepackage{algorithm}
\usepackage{algpseudocodex}

\usepackage{nicefrac}
\usepackage{xspace}
\usepackage{booktabs}
\usepackage{svg}
\usepackage{makecell}
\setsvg{
  inkscapelatex=false
}
\usepackage[inline]{enumitem}
\usepackage{hyperref}
\usepackage[all]{hypcap}
\usepackage[capitalise]{cleveref}
\newlist{questionenum}{enumerate}{1}
\setlist[questionenum]{label=\textbf{(RQ\arabic*)}, ref=(RQ\arabic*)}
\Crefname{question}{Question}{Questions}
\crefname{question}{question}{questions}
\crefalias{questionenumi}{question}

\usepackage{color}

\newcommand{\pow}[1]{\ensuremath{\mathsf{2}^{#1}}\xspace}
\newcommand{\abstractedby}{\in}
\DeclareMathOperator*{\opt}{opt}
\newcommand{\tsp}{\hspace*{1.8mm}}  
\newcommand{\csp}{\hspace*{1mm}}  

\begin{document}
\title{
Adaptive Probabilistic Shielding by \\ Learning MDPs for Safe Reinforcement Learning
}
\titlerunning{Adaptive Probabilistic Shielding by Learning MDPs for Safe RL}
%
\author{
    Astrid~Horn~Brorholt\inst{1}\orcidID{0009-0007-1824-0554} \and
    Maris~F.~L.~Galesloot \inst{2}\orcidID{0009-0002-5112-8584} \and
    Nils~Jansen\inst{2, 3}\orcidID{0000-0003-1318-8973}\and
    Kim~Guldstrand~Larsen\inst{1}\orcidID{0000-0002-5953-3384}\and
    Christian~Schilling\inst{1}\orcidID{0000-0003-3658-1065}
}
\authorrunning{Brorholt et al.}
%
\institute{
    Aalborg University, Aalborg, Denmark \\
    \email{\{asgerhb,kgl,christianms\}@cs.aau.dk} \and
    Radboud University, Nijmegen, Netherlands \\
    \email{maris.galesloot@ru.nl} \and 
    Ruhr University Bochum, Bochum, Germany\\
    \email{n.jansen@rub.de}
}
\maketitle              
\begin{abstract}
Probabilistic shielding is a technique for safe reinforcement learning (RL). 
Typically, a static observer---called the shield---constrains the learning agent's actions to those for which acting safely remains feasible.
Traditionally, the shield is computed from the transition probabilities of the underlying Markov decision process (MDP).
Thus, this technique is not applicable when the MDP model is not given a priori, which, unfortunately, is the case in typical RL applications. 
In this paper, we study the problem of computing a shield in the setting where the transition graph of the MDP is known, but the transition probabilities are unknown. 
Our approach integrates probabilistic shielding with online model learning: as the RL agent explores the environment, we estimate the transition probabilities.
From this estimate, we compute a shield. 
While the shield may be conservative initially, it adapts as the model estimate becomes more precise. 
Thus, the shield improves in tandem with the RL agent. 
This paradigm of \emph{adaptive probabilistic shielding} raises a number of challenges, such as when to recompute the shield and how to balance between exploration and safety during learning. 
We empirically evaluate multiple variants of this paradigm across several environments. 

\end{abstract}

\section{Introduction}
Markov decision processes (MDPs)~\cite{Puterman94} are the standard models to capture \emph{decision-making under uncertainty} in artificial intelligence (AI)~\cite{Kochenderfer15}.
Factors such as unknown or unpredictable environments, contextual changes at runtime, or incomplete data are commonly referred to as \emph{uncertainty}.
Specifically, MDPs capture settings where agents, in each \emph{state} of their environment, choose to execute \emph{actions} upon which the environment \emph{probabilistically transitions} to a new state. 
Upon that transition, the agent receives a \emph{reward}.

Common objectives for MDPs are to (1)~maximize the expected cumulative reward and (2)~adhere to safety constraints specified as temporal logic constraints~\cite{Pnueli77}.
In the past, the first objective was mostly considered by the AI community, and the latter objective by the formal verification community.
Specifically, reinforcement learning (RL) is a major AI technique for decision-making under uncertainty~\cite{DBLP:books/lib/SuttonB2018}.
For an unknown MDP, an RL agent aims to maximize the expected reward by collecting data through exploration of the environment across multiple episodes.
A major limitation in RL is that during exploration, the agent will necessarily execute potentially devastatingly unsafe actions. 
In contrast, probabilistic model checking (PMC) is a formal verification technique that computes the probability of satisfying a safety constraint in an MDP~\cite{BK08}.
The key limitation of PMC is that the MDP must be fully specified. 

\paragraph{Shielded RL.} In response to these key limitations, a tremendous body of work has brought together RL and formal methods in the area of safe RL~\cite{garcia2015comprehensive}, in particular within shielded RL~\cite{DBLP:conf/aaai/AlshiekhBEKNT18,DBLP:conf/atva/DavidJLLLST14}.
Specifically, in \emph{probabilistic pre-shielding}, PMC is used to compute a shield that blocks potentially unsafe actions at runtime~\cite{DBLP:journals/cacm/KonighoferBJJP25,DBLP:conf/concur/0001KJSB20,heckshield2026}.
Such \emph{runtime verification approach} renders RL (more) safe during exploration, yet inherits PMC's strong assumption that the (safety-relevant) environment model, that is, the MDP, must be fully specified.
One may be tempted to approach this problem by first gathering sufficient training data from RL, then using that data to learn a full MDP model of the environment, and finally computing a shield from that model.
However, safety during the data collection is not considered, making such an approach hardly applicable in real-world scenarios.

\paragraph{Problem setting.}
In this paper, we overcome the aforementioned key real-world limitation of shielded RL and propose a practical and adaptive approach.
To that end, we impose mild assumptions about the environment in which the RL agent operates.
First, we assume that simulation access to the true environment MDP is available from the initial state, which is a common assumption in RL.
Second, we assume that the topology, that is, the underlying graph of the MDP, is known, which is much more realistic than knowing the exact transition probabilities.

\begin{figure}[t]
    \centering
     \includegraphics[width=\linewidth,height=46mm,keepaspectratio]{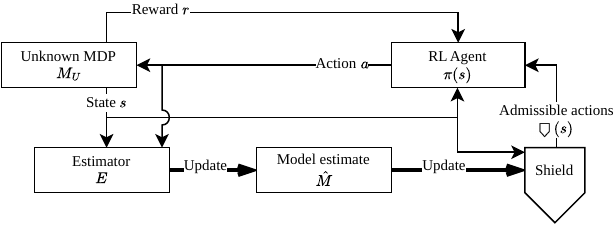}
    \caption{A high-level overview of our adaptive probabilistic shielding approach.}
    \label{fig:flowchart}
\end{figure}

\paragraph{Our approach: Safe RL via Adaptive Probabilistic Shielding.}
\cref{fig:flowchart} shows an overview of our approach.
As usual, the RL agent executes an action in the (unknown) MDP, yielding a reward and causing the environment to transition to a new state.
A key component of our approach is a \emph{model estimator}, based on approaches from~\cite{DBLP:conf/nips/SuilenS0022}.
While the agent interacts with the environment, it collects data on the observed states and actions.
From this data, the model estimator then constructs a \emph{learned} MDP model of the environment.
We use the state-of-the-art PMC tool
PRISM~\cite{PRISM} to compute a shield based on this learned MDP~\cite{galesloot2026robustprobabilisticshieldingsafe}.
At any point during this process, the shield can be updated \emph{adaptively}, and additional data can be collected under the updated shield.

\paragraph{Interval MDPs and shields.}
An essential part of our approach is to implement and compare estimators that yield different types of MDP models from moderate amounts of data collected as a byproduct of the RL process itself. 
In particular, following~\cite{DBLP:conf/nips/SuilenS0022,galesloot2026robustprobabilisticshieldingsafe}, we create so-called interval MDPs (iMDPs)~\cite{DBLP:journals/ior/NilimG05,DBLP:conf/birthday/SuilenBB0025}.
Intuitively, iMDPs capture data uncertainty robustly by defining upper and lower bounds on transition probabilities, based on, for instance, confidence intervals around point estimates of those probabilities. 
Then, PMC can provide upper and lower bounds on the safety probabilities for iMDPs.
In~\cite{galesloot2026robustprobabilisticshieldingsafe}, the worst-case estimates of the bounds are used to compute a \emph{robust}, conservative shield.
Consider the case where a particular action imposes a lower bound of $10\%$ and an upper bound of $30\%$ on the probability of reaching a safety-critical unsafe state.
The application at hand may allow reaching such a state with a maximum probability of $20\%$.
A shield with a \emph{robust} uncertainty interpretation would block that action.

\paragraph{Safety vs.\ exploration.}
The key strength of our adaptive shielding approach is that gathering more data yields more accurate model estimates, which reduce the size of the probability intervals and allow for less conservative shields.
The challenge, however, is that a too-conservative shield may, in the extreme, impede any exploration of the environment and thereby prevent the agent from gathering the necessary data to refine the shield and learn a good policy. 
One solution to this problem is to use an \emph{optimistic} interpretation of uncertainty, as is common in robust RL and referred to as optimism in the face of uncertainty~\cite{DBLP:journals/make/MoosHASCP22}. 
In the example above, the (optimistic) shield would then use the lower probability bound of $10\%$ and allow the critical action.
Another approach is to use common RL exploration techniques to address the exploration-exploitation dilemma~\cite{DBLP:books/lib/SuttonB2018}. 
 
\paragraph{Contributions and research questions.}
The main contribution of this paper is a novel, adaptive shielding algorithm that accounts for the uncertainty in estimating an MDP from data. 
We provide a thorough experimental evaluation structured around several concrete research questions.
First, we evaluate if an adaptive shielding approach is beneficial to (1) obtain a safe and reward-optimal policy after training and (2) remain safe during training. 
Then, we investigate whether the choice of model estimator affects the performance of the shield and how far the quality of the MDP estimate improves over time with respect to the (conservativeness) of the shield.
Finally, we take into account various practical considerations, such as the number of model updates in relation to the number of RL episodes. 
The paper is structured as follows. 
In the remainder of the introduction, we discuss related work. 
In \cref{sec:prelims}, we provide necessary background, followed by the definitions of shields for (interval) MDPs and model estimation in \cref{sec:shielding-and-estimation}.
\cref{sec:adaptive-shields} describes our adaptive probabilistic shielding algorithm.
Finally, we present our research questions and experimental analysis in \cref{sec:evaluation}.

\subsection{Related work}

\paragraph{Probabilistic shielding.}
Classic shields provide unconditional safety guarantees, but this is often too conservative~\cite{heckshield2026,DBLP:journals/cacm/KonighoferBJJP25}.
Shielding has been extended to many classes of models~\cite{DBLP:conf/atal/Elsayed-AlyBAET21,DBLP:conf/vecos/BrorholtJLLS23,DBLP:journals/corr/abs-2509-12085,DBLP:conf/l4dc/KimCRLPBSF25,DBLP:conf/aaai/AlshiekhBEKNT18,DBLP:conf/amcc/ReedL25,DBLP:journals/corr/abs-2510-03481,BrorholtLS25}.
We consider probabilistic shields that permit a level of risk of reaching unsafe states~\cite{DBLP:conf/aaai/CourtBG25,DBLP:conf/concur/0001KJSB20,heckshield2026,DBLP:journals/cacm/KonighoferBJJP25}.
In this paper, we focus on practical shields that guarantee the \emph{admissibility} of probabilistic safety guarantees~\cite{DBLP:conf/concur/0001KJSB20,DBLP:conf/amcc/PrangerKTD0B21,galesloot2026robustprobabilisticshieldingsafe,DBLP:journals/cacm/KonighoferBJJP25}.
Specifically, our shielding method is based on~\cite{galesloot2026robustprobabilisticshieldingsafe}.

\paragraph{Adaptive shielding.}
Several works have considered shields that change over time.
Pranger et al.\ adaptively construct a finite-state environment abstraction from past observations to maintain a shield~\cite{DBLP:conf/amcc/PrangerKTD0B21}.
Similarly, Tappler et al.\ present an iterative approach using automata learning~\cite{DBLP:conf/isola/TapplerPKMBL22}.
In other work, the shield adapts to changes in the environment, assuming white-box access to the (parametric) dynamics~\cite{senthilvelan_similarity-based_2023,DBLP:journals/pacmpl/FengZPL25}.
Goodall et al.\ estimate probabilistic safety by simulating future trajectories in a learned latent model~\cite{DBLP:conf/ecai/GoodallB23}.
Bethell et al.\ learn a shield using an auto-encoder and adjust the safety threshold in a subsequent RL phase~\cite{DBLP:conf/ecai/BethellGCI25}.

\paragraph{Shielding based on model estimates.}
Galesloot et al.\ estimate iMDPs for shielding in \emph{offline} RL~\cite{galesloot2026robustprobabilisticshieldingsafe}.
Suilen et al.\ obtain iMDP estimates via optimistic exploration and then analytically compute a robust policy~\cite {DBLP:conf/nips/SuilenS0022}.
Another recent work collects environment data offline and then computes a robust shield for shielded RL, with the main contribution being a new algorithm to compute the shield~\cite{court2026robustshieldingsafereinforcement}.

\paragraph{Delimitation.}
The main differences of our problem setting are as follows. We assume a static environment with a known, finite state space and transition structure, but with unknown transition probabilities.
We further assume black-box access instead of a settable simulator, which is why we specifically care about safety during the whole process, including data collection and exploration.
Our method differs from previous work in that we integrate model estimation and updates of both the shield and the policy into a single integrated \emph{online} RL procedure with adaptive probabilistic shields.

\section{Preliminaries}\label{sec:prelims}

\paragraph{Probability distributions.} Given a set~$X$, a probability distribution~$p \colon X \rightarrow [0, 1]$ satisfies $\sum_{x \in X} p(x) = 1$.
Let $\Delta(X)$ denote the set of probability distributions over $X$ and let~$\mathrm{Unif}_X \in \Delta(X)$ denote the uniform distribution over~$X$.

\paragraph{Intervals.} A closed interval $[a, b] \subseteq \mathbb{R}$ for $a \leq b$ describes the set $\{x \in \mathbb{R} \mid a \leq x \leq b \}$.
An open interval $(a, b)$ describes $\{x \in \mathbb{R} \mid a < x < b \}$. Half-open intervals are defined analogously.
Let $\mathbb{I} = \{ [a, b] \mid 0 < a \leq b \leq 1 \}$ denote the set of uncertain nonzero probabilities.

\paragraph{Markov decision processes.}
A \emph{Markov decision process} (MDP) is a tuple~$M = (S, A, s_0, T)$ where~$S$ is the finite set of states, $A$ is the finite set of actions, $s_0 \in S$ is the initial state, and $T \colon S \times A \times S \to [0,1]$ is the probabilistic transition function satisfying $\sum_{s' \in S} T(s, a, s') = 1$ for all~$s$ and~$a$.
A \emph{run} is an alternating sequence~$s_0 a_0 s_1 a_1 \dots$ of states and actions such that~$T(s_i, a_i, s_{i+1}) > 0$ for all~$i$.

We consider two types of policies.
A \emph{deterministic policy}~$\pi \colon S \to A$ maps each state to an action.
A \emph{nondeterministic policy}~$\pi_N \colon S \to \pow{A}$ maps each state to a set of actions.
A run~$s_0 a_0 s_1 a_1 \dots$ is an \emph{outcome} of a deterministic policy~$\pi$ (resp.\ nondeterministic policy~$\pi_N$) if~$a_i = \pi(s_i)$ (resp.~$a_i \in \pi_N(s_i)$) for all~$i$.

An \emph{unknown} MDP (uMDP) is a tuple~$M_U = (S, A, s_0, T_U)$ where~$S$, $A$, and~$s_0$ are defined as for MDPs and~$T_U \colon S \times A \to \pow{S}$ is a nondeterministic transition function (i.e., uMDPs are ordinary transition systems).
Each MDP induces an unknown MDP by removing impossible transitions and dropping the probabilities; formally: $T_U(s,a) = \{s' \in S \mid T(s, a,s') > 0\}$.
An \emph{interval} MDP (iMDP)~\cite{DBLP:conf/lics/JonssonL91,DBLP:journals/ior/NilimG05,DBLP:journals/jcss/StrehlL08,DBLP:conf/isola/Jaeger0BLJ20,DBLP:conf/nips/SuilenS0022,DBLP:conf/birthday/SuilenBB0025} is a tuple $M_I = (S, A, s_0, T_I)$ where again~$S$, $A$, and $s_0$ are defined as for MDPs and~$T_I \colon S \times A \times S \to \mathbb{I} \cup \{0\}$ is an \emph{interval transition function}.
Consider an MDP $M = (S, A, s_0, T)$ and an iMDP $M_I = (S, A, s_0, T_I)$ over~$S$ and~$A$.
We say that~$T_I$ \emph{abstracts}~$T$, written $T \abstractedby T_I$, if $T$ is a probabilistic transition function and each interval in~$T_I$ contains the corresponding probability in~$T$; formally:
$\forall s \in S ~ \forall a \in A \colon \sum_{s' \in S} T(s, a, s') = 1
\land
\forall s' \in S \colon T(s, a, s') \in T_I(s, a, s')$.
Analogously, we say that~$M_I$ abstracts~$M$, written~$M \abstractedby M_I$.
Note that, because intervals in~$\mathbb{I}$ must not include~$0$, if~$T \abstractedby T_I$ and~$T' \abstractedby T_I$, then the same transitions in~$T$ and~$T'$ have a non-zero probability; formally: $\forall T, T' \abstractedby T_I ~\forall s, s' \in S ~\forall a \in A \colon T(s, a, s') > 0 \implies T'(s, a, s') > 0$.

\paragraph{Reinforcement learning.}
We assume that the reader is familiar with reinforcement learning (RL)~\cite{DBLP:books/lib/SuttonB2018} and only recall some basic commonalities.
Let $R \colon S \times A \times S \to \mathbb{R}$ be the reward function and $\gamma \in [0,1)$ a discount factor.
Given a deterministic policy $\pi$, the \emph{expected cumulative discounted reward} from the initial state~$s_0$ is
$
V^\pi(s_0)
=
\mathbb{E}^{\pi}_{s}
\left[
\sum_{t=0}^{\infty}
\gamma^t
R(s_t,a_t,s_{t+1})
\right],
$
where at each step $t\in\mathbb{N}$, the action is $a_t = \pi(s_t)$ and the expectation is taken over the probabilistic state transitions governed by the transition function~$T$.
In RL, an agent explores an environment (which is assumed to be an MDP) with the aim to learn a policy that maximizes the expected cumulative discounted reward
from the exploration experience.
This ``training'' takes place in episodes of multiple steps each.
RL requires only black-box sampling access to the environment MDP from an initial state.
Two major paradigms are \emph{model-free} and \emph{model-based} RL~\cite{DBLP:books/lib/SuttonB2018}.
The latter learns an approximation of the MDP as the agent explores.
On the one hand, the algorithm proposed in this paper learns such an approximation to construct the shield, and is therefore model-based.
On the other hand, the algorithm also includes an RL component, for which our prototype implementation uses Q-learning (which is model-free).
We note that most other RL algorithms (model-free or model-based) could also be used in place of Q-learning.

During training, we use an \emph{$\varepsilon$-greedy exploration/exploitation} strategy~\cite{DBLP:books/lib/SuttonB2018}: at each step, the agent chooses a random action with probability~$\varepsilon$ (``exploration'') and follows the (partially) learned policy with probability~$1 - \varepsilon$ (``exploitation'').

\paragraph{Safety and shielding.}
We consider safety properties~$\varphi \subseteq S$ given as a set of safe states.
A run~$s_0 a_0 s_1 a_1 \dots$ is \emph{safe} if~$s_i \in \varphi$ for all~$i$.

Given a deterministic policy~$\pi \colon S \to A$, a \emph{shield} is any nondeterministic policy $\ensuremath{\text{\protect 
{\color{white} $\nabla$} 
\hspace{-1.5em}
\tikz[baseline]{
    \draw [line width=0.05em]
        (0, 0.5em) -- 
        (0.45em, 0.5em) --
        (0.45em, 0.13333em) --
        (0.225em, -0.1em) --
        (0, 0.13333em) --
        cycle
}
\hspace{-0.65em}}}\xspace \colon S \to \pow{A}$ over the same states and actions.
The \emph{shielded policy}~$\pi_{\ensuremath{\text{\protect 
{\color{white} $\nabla$} 
\hspace{-1.5em}
\tikz[baseline]{
    \draw [line width=0.05em]
        (0, 0.5em) -- 
        (0.45em, 0.5em) --
        (0.45em, 0.13333em) --
        (0.225em, -0.1em) --
        (0, 0.13333em) --
        cycle
}
\hspace{-0.65em}}}\xspace}$ is a deterministic policy that acts similarly to~$\pi$ but only chooses actions allowed by the shield, i.e., $\forall s \in S ~\forall a \in A \colon \pi_{\ensuremath{\text{\protect 
{\color{white} $\nabla$} 
\hspace{-1.5em}
\tikz[baseline]{
    \draw [line width=0.05em]
        (0, 0.5em) -- 
        (0.45em, 0.5em) --
        (0.45em, 0.13333em) --
        (0.225em, -0.1em) --
        (0, 0.13333em) --
        cycle
}
\hspace{-0.65em}}}\xspace}(s) = a \implies a \in \ensuremath{\text{\protect 
{\color{white} $\nabla$} 
\hspace{-1.5em}
\tikz[baseline]{
    \draw [line width=0.05em]
        (0, 0.5em) -- 
        (0.45em, 0.5em) --
        (0.45em, 0.13333em) --
        (0.225em, -0.1em) --
        (0, 0.13333em) --
        cycle
}
\hspace{-0.65em}}}\xspace(s)$.
Additionally, the shield only alters actions when necessary: $\forall s \in S \colon \pi(s) \in \ensuremath{\text{\protect 
{\color{white} $\nabla$} 
\hspace{-1.5em}
\tikz[baseline]{
    \draw [line width=0.05em]
        (0, 0.5em) -- 
        (0.45em, 0.5em) --
        (0.45em, 0.13333em) --
        (0.225em, -0.1em) --
        (0, 0.13333em) --
        cycle
}
\hspace{-0.65em}}}\xspace(s) \implies \pi_\ensuremath{\text{\protect 
{\color{white} $\nabla$} 
\hspace{-1.5em}
\tikz[baseline]{
    \draw [line width=0.05em]
        (0, 0.5em) -- 
        (0.45em, 0.5em) --
        (0.45em, 0.13333em) --
        (0.225em, -0.1em) --
        (0, 0.13333em) --
        cycle
}
\hspace{-0.65em}}}\xspace (s) = \pi(s)$.
The action chosen by~$\pi_\ensuremath{\text{\protect 
{\color{white} $\nabla$} 
\hspace{-1.5em}
\tikz[baseline]{
    \draw [line width=0.05em]
        (0, 0.5em) -- 
        (0.45em, 0.5em) --
        (0.45em, 0.13333em) --
        (0.225em, -0.1em) --
        (0, 0.13333em) --
        cycle
}
\hspace{-0.65em}}}\xspace(s)$ when $\pi(s) \notin \ensuremath{\text{\protect 
{\color{white} $\nabla$} 
\hspace{-1.5em}
\tikz[baseline]{
    \draw [line width=0.05em]
        (0, 0.5em) -- 
        (0.45em, 0.5em) --
        (0.45em, 0.13333em) --
        (0.225em, -0.1em) --
        (0, 0.13333em) --
        cycle
}
\hspace{-0.65em}}}\xspace(s)$ depends on the implementation of the RL agent.
In our implementation, we use Q-learning~\cite{DBLP:journals/ml/WatkinsD92}, for which it is straightforward to read out the best admissible action in~$\ensuremath{\text{\protect 
{\color{white} $\nabla$} 
\hspace{-1.5em}
\tikz[baseline]{
    \draw [line width=0.05em]
        (0, 0.5em) -- 
        (0.45em, 0.5em) --
        (0.45em, 0.13333em) --
        (0.225em, -0.1em) --
        (0, 0.13333em) --
        cycle
}
\hspace{-0.65em}}}\xspace(s)$.

\section{Probabilistic Shielding Using an Estimator}\label{sec:shielding-and-estimation}

In this paper, we construct shields based on the method described in~\cite{galesloot2026robustprobabilisticshieldingsafe}, which was originally developed for the \emph{offline} RL problem where a fixed dataset is given.
Our approach differs in that we continuously adapt the shield based on newly generated data, and we must consider exploration to collect new data within and beyond
the shield's allowed actions.
As we will see later, such a setting is particularly challenging and yields trade-offs between safety and exploration.

\subsection{Probabilistic shielding approaches for (interval) MDPs}
\label{sec:shields}
Next, we recall how to obtain a shield~$\ensuremath{\text{\protect 
{\color{white} $\nabla$} 
\hspace{-1.5em}
\tikz[baseline]{
    \draw [line width=0.05em]
        (0, 0.5em) -- 
        (0.45em, 0.5em) --
        (0.45em, 0.13333em) --
        (0.225em, -0.1em) --
        (0, 0.13333em) --
        cycle
}
\hspace{-0.65em}}}\xspace$ for an (interval) MDP~$M$ and a safety specification~$\varphi$.
We assume a horizon~$h \in \mathbb{N}$ and parameters $\theta, \kappa \in [0, 1]$, which we explain below.
Before the formalization, we first describe the high-level idea.
Intuitively, the shield ensures the existence of a series of~$h$ actions from the current state~$s$ such that the chance of a safety violation along these steps is below~$\theta$.
For tractability, the shield is memoryless and thus ignores accumulated past risk before reaching the current state~$s$.
Since the restriction is probabilistic and memoryless, there may still be a chance of reaching a state where this guarantee does not hold~\cite{galesloot2026robustprobabilisticshieldingsafe,DBLP:journals/cacm/KonighoferBJJP25,heckshield2026}.
Whenever no sufficiently safe action is available, the shield only allows actions that are~$\kappa$-close to the safest available action.

Now we formalize this idea.
Let $\varphi|{h}$ be the set of all \emph{$h$-safe runs} $s_0 a_0 s_1 a_1 \dots$ with safe $h$-prefix, i.e., $s_i \in \varphi$ for~$i \leq h$.
Let~$\mathbb P_\pi^{M, \varphi|{h}}(s)$ be the probability of an MDP~$M$ producing an $h$-safe run by following policy~$\pi$ starting in state~$s$.
We denote the related safety optimization problem by
$\mathbb{P}_{\max}^{M, \varphi|h} (s) = \max_\pi \mathbb{P}_\pi^{M, \varphi|h} (s)$.
Moreover, let the probability of producing a run in~$\varphi|h$ after taking action~$a$ in state~$s$ be
$\mathbb{P}_{\max}^{M, \varphi|h} (s, a) = \sum_{s' \in S} T(s, a, s') \mathbb{P}_{\max}^{M, \varphi|h-1} (s')$.

Normally, the shield allows all actions guaranteeing a safe $h$-step run with probability at least~$1 - \theta$; formally: $\ensuremath{\text{\protect 
{\color{white} $\nabla$} 
\hspace{-1.5em}
\tikz[baseline]{
    \draw [line width=0.05em]
        (0, 0.5em) -- 
        (0.45em, 0.5em) --
        (0.45em, 0.13333em) --
        (0.225em, -0.1em) --
        (0, 0.13333em) --
        cycle
}
\hspace{-0.65em}}}\xspace_\theta(s) = \{ a \in A \mid \mathbb{P}_{\max}^{M, \varphi|h} (s, a) \geq 1 - \theta \}$.
However, a shielded policy may still reach a state~$s$ where this shield definition would not allow any action (i.e., $\ensuremath{\text{\protect 
{\color{white} $\nabla$} 
\hspace{-1.5em}
\tikz[baseline]{
    \draw [line width=0.05em]
        (0, 0.5em) -- 
        (0.45em, 0.5em) --
        (0.45em, 0.13333em) --
        (0.225em, -0.1em) --
        (0, 0.13333em) --
        cycle
}
\hspace{-0.65em}}}\xspace_\theta(s) = \emptyset$).
In that case, the shield instead allows all actions that are $\kappa$-close to the safest available action;
formally:
$\ensuremath{\text{\protect 
{\color{white} $\nabla$} 
\hspace{-1.5em}
\tikz[baseline]{
    \draw [line width=0.05em]
        (0, 0.5em) -- 
        (0.45em, 0.5em) --
        (0.45em, 0.13333em) --
        (0.225em, -0.1em) --
        (0, 0.13333em) --
        cycle
}
\hspace{-0.65em}}}\xspace_\kappa(s) = \{ a \in A \mid \mathbb{P}_{\max}^{M, \varphi|h} (s, a) \geq \max_{a'} \mathbb{P}_{\max}^{M, \varphi|h} (s, a') - \kappa \}$.
Our shield combines these two cases:
\begin{equation}
    \ensuremath{\text{\protect 
{\color{white} $\nabla$} 
\hspace{-1.5em}
\tikz[baseline]{
    \draw [line width=0.05em]
        (0, 0.5em) -- 
        (0.45em, 0.5em) --
        (0.45em, 0.13333em) --
        (0.225em, -0.1em) --
        (0, 0.13333em) --
        cycle
}
\hspace{-0.65em}}}\xspace(s) = \begin{cases}
        \ensuremath{\text{\protect 
{\color{white} $\nabla$} 
\hspace{-1.5em}
\tikz[baseline]{
    \draw [line width=0.05em]
        (0, 0.5em) -- 
        (0.45em, 0.5em) --
        (0.45em, 0.13333em) --
        (0.225em, -0.1em) --
        (0, 0.13333em) --
        cycle
}
\hspace{-0.65em}}}\xspace_\theta (s)
        & \mathrm{if} ~ \ensuremath{\text{\protect 
{\color{white} $\nabla$} 
\hspace{-1.5em}
\tikz[baseline]{
    \draw [line width=0.05em]
        (0, 0.5em) -- 
        (0.45em, 0.5em) --
        (0.45em, 0.13333em) --
        (0.225em, -0.1em) --
        (0, 0.13333em) --
        cycle
}
\hspace{-0.65em}}}\xspace_\theta(s) \neq \emptyset
        \\
        \ensuremath{\text{\protect 
{\color{white} $\nabla$} 
\hspace{-1.5em}
\tikz[baseline]{
    \draw [line width=0.05em]
        (0, 0.5em) -- 
        (0.45em, 0.5em) --
        (0.45em, 0.13333em) --
        (0.225em, -0.1em) --
        (0, 0.13333em) --
        cycle
}
\hspace{-0.65em}}}\xspace_\kappa (s)
        & \mathrm{otherwise.}
    \end{cases}
    \label{eq:shield}
\end{equation}

Following~\cite{galesloot2026robustprobabilisticshieldingsafe},
we extend shields to iMDPs by defining the probability of producing a run for iMDPs as follows.
Let
$\mathbb{P}_{\max\opt}^{M_I, \varphi|h} (s) = \max_\pi \opt_{M \in M_I} \mathbb{P}_\pi^{M, \varphi|h} (s)$ be the probability of producing a run under the model $M \in M_I$ corresponding to the optimization direction $\opt \in \{\min,\max\}$.
Moreover, let the iMDP version of the probability producing a run in~$\varphi|h$ after taking action~$a$ in state~$s$ be
\begin{align}
\label{eq:imdpshield}
\mathbb{P}_{\max\opt}^{M_I, \varphi|h} (s, a) 
&= \opt_{{T}^{\dagger} \in T_I} \, \sum_{s' \in S} T^{\dagger}(s, a, s') \mathbb{P}_{\max\opt}^{M_I, \varphi|h-1} (s').
\end{align}
We use the probabilistic model checker PRISM~\cite{PRISM} to efficiently compute such probabilities on iMDPs.
Statistically speaking, a worst-case assumption ($\opt=\min$) makes the shield \emph{robust} against estimation errors.
Following~\cite{galesloot2026robustprobabilisticshieldingsafe}, we define a (pessimistic) \emph{robust shield} for an iMDP~$M_I$ similarly to \cref{eq:shield} using $\mathbb{P}_{\max\min}^{M_I, \varphi|h} (s, a) $.
Instead of assuming the worst-case MDP~$M$ from an estimate~$M_I$,
one can also assume the best case $(\opt=\max)$,
specified as 
$\mathbb{P}_{\max \max}^{M_I, \varphi|h} (s)$.
We call this alternative an \emph{optimistic shield}.
We say that the \emph{attitude} of a shield is either robust or optimistic, depending on how it was constructed from an iMDP.

\subsection{Estimators from data for unknown MDPs}
\label{sec:introduce:estimators}
We recall three existing estimators, based on those that appeared in \cite{DBLP:conf/nips/SuilenS0022}.
While exploring the black-box MDP, we count how many times a transition triple $(s, a, s')$ has been observed.
For that, we use a transition database $D \colon S \times A \times S \to \mathbb{N}$.
Let $D(s,a) = \sum_{s'} D(s,a,s')$ be the \emph{total count} for a state-action pair $(s,a)$.

An \emph{estimator} is a function $E(M_U, D)$ that maps a uMDP~$M_U$ and a transition database~$D$ to either an estimated MDP~$\hat M$ or iMDP~$\hat{M}_I$, depending on the estimator.
Below, we describe three estimators that learn (i.e., estimate the transition function of) MDPs or iMDPs, with or without guarantees: \emph{MAP} ($E_{\textrm{MAP}}$), \emph{PAC} ($E_{\textrm{PAC}}$), and \emph{LUI} ($E_{\textrm{LUI}}$).
These approaches estimate the transition function locally for each $(s,a)$-pair using knowledge of the graph in the form of a given uMDP $M_U$, where $T_U(s,a)$ denotes the set of successor states.

\paragraph{MAP.}
The first approach finds a \emph{point estimate} $\hat{T}$ of the MDP's transition function~$T$ based on the data $D$.
Following~\cite{DBLP:conf/nips/SuilenS0022}, we define point estimates as maximum a-posteriori (MAP) estimation with respect to a symmetric prior weight assigned to each successor state, which we denote as a single $w \in \mathbb{N}$.
Then 
\[
\hat{T}(s,a,s') = \frac{w + D(s,a,s')-1}{\left(\sum_{t \in T_U(s,a)} w + D(s,a,t)\right) - |T_U(s,a)|}
\]
defines the MAP point estimate.
It can be viewed as a maximum-likelihood probability with some additive smoothing from $w$.
The MAP estimator $E_{\textrm{MAP}}$ maps $(M_U, D)$ to an estimated MDP $\hat{M}$ with point estimates $\hat{T}$.

\paragraph{PAC.}
The point estimates of the MAP estimator do not account for the uncertainty arising from estimating probabilities from data.
Point estimates can be turned into \emph{probably approximately correct}~(PAC) intervals via Hoeffding's inequality~\cite{Hoeffding}, such that the estimated iMDP contains the true MDP with high probability $1 - \delta$, for $\delta \in [0,1]$~\cite{DBLP:conf/cav/AshokKW19,DBLP:conf/nips/SuilenS0022,galesloot2026robustprobabilisticshieldingsafe}.
As such, by the union bound, we distribute $\delta$ over all transitions as $\delta_{T} = \nicefrac{\delta}{\sum_{s,a} k(s,a)}$, where $k(s,a)$ denotes the number of successor states $k(s,a)=|T_U(s,a)|$ if $|T_U(s,a)| > 1$ and $k(s,a) = 0$ otherwise. 
Then, $\eta_{s,a} = \nicefrac{\log [\nicefrac{2}{\delta_T}]}{2\cdot D(s,a)}$ denotes the range of the PAC interval around the point estimate for $(s,a)$.
Using each $\eta_{s,a}$, we construct the intervals
\begin{equation}
\label{eq:pac}
\hat{T}_{I}(s,a,s') = \left[\max(\xi, \hat{T}(s,a,s') - \eta_{s,a}), \min(1, \hat{T}(s,a,s') + \eta_{s,a}) \right]
\end{equation}
where $\xi \in (0,1)$ is a small constant that ensures intervals for transitions with nonzero probability (as given by the unknown MDP $M_U$) map to $[\xi, 1]$.
The PAC estimator $E_{\textrm{PAC}}$ maps $(M_U, D)$ to an iMDP $\hat{M}_I$ with intervals $\hat{T}_I$ from \cref{eq:pac}.

\emph{LUI.}
The third approach that we consider is the \emph{linearly updating intervals}~(LUI) estimator from~\cite{DBLP:conf/nips/SuilenS0022}, which in turn is based on~\cite{imprecisionconflicts2009}.
While it does not retain PAC guarantees, it iteratively learns probabilities by updating intervals.
We assign each unknown transition a prior interval $\tilde{T}_I(s,a,s_i') = [\underline{T}_i, \overline{T}_i]$ and prior strength $[\underline{n}_i,\overline{n}_i]$.
The strength influences the prior's effect on the updated intervals.
At any point, we find new intervals given the database $D$ by distinguishing cases based on whether the current intervals agree with the new data.
For any $(s,a)$ and $s'_j$, let $F_j = \nicefrac{D(s,a,s'_j)}{D(s,a)}$ denote the relative occurrence of transition $(s,a,s'_j)$ in the database~$D$. 
Then, the updates are:

\begin{equation}
\label{eq:lui}
\underline{T}_i \gets 
    \begin{cases}
        \vspace{0.5em}
       \frac{\overline{n}_i \underline{T}_i + D(s,a,s'_i)}{\overline{n}_i + D(s,a)} \quad \text{if } F_j \geq \underline{T}_j \text{ for all } s_j',\\
        \frac{\underline{n}_i \underline{T}_i + D(s,a,s'_i)}{\underline{n}_i + D(s,a)} \quad \text{otherwise}.
    \end{cases}
\end{equation}
\begin{equation}
\overline{T}_i \gets 
    \begin{cases}
        \vspace{0.5em}
       \frac{\overline{n}_i \overline{T}_i + D(s,a,s'_i)}{\overline{n}_i + D(s,a)} \quad \text{if } F_j \leq \overline{T}_j \text{ for all } s_j',\\
        \frac{\underline{n}_i \overline{T}_i + D(s,a,s'_i)}{\underline{n}_i + D(s,a)} \quad \text{otherwise}.
    \end{cases}
\end{equation}

The (strength) intervals are found from total counts $[\underline{n}_i + D(s,a),\overline{n}_i + D(s,a)]$.
Initial intervals are valid when $0 < \underline{T}_i \leq \overline{T}_j \leq 1$ and $\overline{n}_i \geq \underline{n}_i \geq 1$. 
Similarly to $E_{\textrm{PAC}}$, $E_{\textrm{LUI}}$ maps $(M_U, D) $ to an iMDP $\tilde{M}_I$ with intervals $\tilde{T}_I$ using \cref{eq:lui}.

\paragraph{Guarantees and convergence.}
While only $E_{\textrm{PAC}}$ provides statistical guarantees from finite data~\cite{DBLP:conf/cav/AshokKW19}, all three estimators converge to the true probabilities as the number of visits to each transition tends to infinity~\cite{DBLP:conf/nips/SuilenS0022}.

\section{Adaptive Probabilistic Shielding}\label{sec:adaptive-shields}

In this section, we develop the paradigm that we call \emph{adaptive probabilistic shielding}.
Before we present our algorithm, we motivate the problem it addresses.

\subsection{Problem Statement}\label{sec:problem}

We consider an RL application in a safety-critical real-world scenario.
In particular, we do not know the underlying MDP model (only the underlying uMDP) and hence do not assume access to a settable simulator.
While safety violations may not be entirely avoidable, we place great importance on them, as they may occur during real-world data collection outside a simulator~\cite{DBLP:journals/ijrr/LacerdaFPH19}.
Hence, our goal is to obtain a policy~$\pi$ subject to three sub-goals: (i)~achieve a given admissibility threshold of the safety specification,
(ii)~achieve a high expected reward, and (iii)~achieve a low number of safety violations during training.

To highlight the intricacy of our problem, we point out that goal~(iii) is in direct competition with the other two goals.
This is because higher safety during training requires more conservative exploration, which may prevent the discovery of a better-performing policy (e.g., a faster and/or safer route to a goal state).
Since we do not assume prior knowledge of the environment's transition dynamics, one may have to take more risks to learn them; thus, an action deemed less safe due to higher uncertainty may only be determined to be safer after enough exploration.
While RL solves~(ii), it does not achieve~(i), and it may also perform poorly regarding~(iii), since it typically relies on (random) exploration of the environment.
To additionally achieve~(i), we could apply shielded RL; however, since we do not know the MDP, we would need to learn an MDP or iMDP model, which would itself require exploration for the data collection and thus again fail to achieve goal~(iii).

\subsection{Adaptive Probabilistic Shielding}\label{sec:ourapproach}

Our answer to this dilemma is to interweave all three procedures (policy learning, shield construction, and model estimation) into a single adaptive learning loop.
Generally, we collect data from the RL agent's exploration of the environment.
From time to time, we use that data to update our model estimate, and from that improved estimate, we obtain a refined shield that allows us to continue exploring the environment more safely and/or less conservatively.

One may be tempted to think that this process will, given enough episodes, converge to the ideal solution of learning the underlying MDP and thus the best possible shield.
However, this is not necessarily the case, and indeed, we observed that such an approach can fail in practice.
The issue is that the conservative shield, from the beginning, may simply prevent exploration of large parts of the state space, even if the corresponding actions were to be perfectly safe under the true MDP model.
More specifically, the shield cannot distinguish between actions that are already known to be risky (which it should indeed block) and actions for which the model estimate is too coarse to make a definite judgement.
This is particularly pronounced for the more pessimistic robust shields.

Our final step is to extend the $\varepsilon$-greedy exploration strategy to explore beyond the shield's boundaries.
When the exploration strategy decides to explore a random action in state~$s$, we choose this action from the full set of actions~$A$, rather than just from~$\ensuremath{\text{\protect 
{\color{white} $\nabla$} 
\hspace{-1.5em}
\tikz[baseline]{
    \draw [line width=0.05em]
        (0, 0.5em) -- 
        (0.45em, 0.5em) --
        (0.45em, 0.13333em) --
        (0.225em, -0.1em) --
        (0, 0.13333em) --
        cycle
}
\hspace{-0.65em}}}\xspace(s)$ allowed by the shield.
We show the impact of this extension empirically in the next section in~\ref{sec:unshieldedexploration}.

\begin{algorithm}[t]
    \caption{Safe RL via Adaptive Probabilistic Shielding}
    \begin{algorithmic}[1]
        \Require{%
            black-box MDP~$M = (S, A, s_0, T)$;
            uMDP~$M_U = (S, A, s_0, T_U)$;
            reward function~$R$;
            safety specification~$\varphi$;
            estimator~$E$;
            shield update delay~$u \in \mathbb{N}$;
            shield parameters~$\theta, \kappa \in [0, 1]$ and $h \in \mathbb{N}$;
            exploration rate~$\varepsilon \in [0, 1]$;
            number of episodes~$N \in \mathbb{N}$;
            maximum episode length~$L \in  \mathbb{N}$
        }
        \\[-.5\baselineskip]
        \State $\pi \gets$ Initialize RL policy
        \State $D(s,a,s') \gets 0, \quad \forall \, (s,a,s')$ \Comment{Initialize transition database}
        \For {$i$ \textbf{in}~$0$ \textbf{to}~$N-1$}\label{ln:outerloop}
            \If {$i \equiv 0 \mod u$} \label{ln:shieldrecomputationconditional}
                \State $\hat{M} \gets E(M_U, D)$ \Comment{See \Cref{sec:introduce:estimators}}
                \State $\ensuremath{\text{\protect 
{\color{white} $\nabla$} 
\hspace{-1.5em}
\tikz[baseline]{
    \draw [line width=0.05em]
        (0, 0.5em) -- 
        (0.45em, 0.5em) --
        (0.45em, 0.13333em) --
        (0.225em, -0.1em) --
        (0, 0.13333em) --
        cycle
}
\hspace{-0.65em}}}\xspace \gets$ Synthesize shield from $\hat{M}$, $\theta$, $\kappa$, and $h$ for $\varphi$ \Comment{See \Cref{sec:shields}}
            \EndIf
            \State $s \gets s_0$
            \For {$j$ \textbf{in}~$0$ \textbf{to}~$L-1$} \label{ln:innerloop}
                \If {Random bit with probability $\varepsilon$ of being true}
                    \State $a \sim \text{Unif}_A$ \Comment{``Explore'' -- uniform choice among all actions} \label{ln:explore}
                \Else
                    \State $a = \pi_{\ensuremath{\text{\protect 
{\color{white} $\nabla$} 
\hspace{-1.5em}
\tikz[baseline]{
    \draw [line width=0.05em]
        (0, 0.5em) -- 
        (0.45em, 0.5em) --
        (0.45em, 0.13333em) --
        (0.225em, -0.1em) --
        (0, 0.13333em) --
        cycle
}
\hspace{-0.65em}}}\xspace}(s)$ \Comment{``Exploit'' -- agent chooses the best safe action} \label{ln:exploit}
                \EndIf
                \State $s' \sim T(s, a)$ \Comment{Take a step with action $a$ in the environment}
                \State $\pi \gets$ Update policy with transition $(s, a, s')$ and reward $R(s, a, s')$
                \State $D(s,a,s') \gets D(s,a,s') + 1$
                \State $s \gets s'$
            \EndFor
        \EndFor
        \Return{($\pi, \ensuremath{\text{\protect 
{\color{white} $\nabla$} 
\hspace{-1.5em}
\tikz[baseline]{
    \draw [line width=0.05em]
        (0, 0.5em) -- 
        (0.45em, 0.5em) --
        (0.45em, 0.13333em) --
        (0.225em, -0.1em) --
        (0, 0.13333em) --
        cycle
}
\hspace{-0.65em}}}\xspace, \hat{M}$) \phantom{\huge X}} 
    \end{algorithmic}
    \label{alg:alg}
\end{algorithm}

\cref{alg:alg} shows the pseudocode of our proposed approach.
Notably, we only use white-box access to the environment via the uMDP~$M_U$, while we only access the underlying MDP~$M$ implicitly via the black-box functions~$T$ and~$R$.

In our implementation, we use Q-learning~\cite{DBLP:journals/ml/WatkinsD92} to find a policy~$\pi$.
For that, we initialize~$\pi$ as an empty Q-table and an empty database~$D$ of observed transition triples.
In the first iteration of the outer for-loop in Line~\ref{ln:outerloop}, $i=0$ satisfies the condition in Line~\ref{ln:shieldrecomputationconditional}, which triggers the computation of the first model estimate~$\hat{M}$ and shield~$\ensuremath{\text{\protect 
{\color{white} $\nabla$} 
\hspace{-1.5em}
\tikz[baseline]{
    \draw [line width=0.05em]
        (0, 0.5em) -- 
        (0.45em, 0.5em) --
        (0.45em, 0.13333em) --
        (0.225em, -0.1em) --
        (0, 0.13333em) --
        cycle
}
\hspace{-0.65em}}}\xspace$.
Since the transition database is still empty, this estimate and shield are most conservative.
The inner for-loop in Line~\ref{ln:innerloop} represents Q-learning for a single episode with the extended $\varepsilon$-greedy exploration strategy described above under~$\ensuremath{\text{\protect 
{\color{white} $\nabla$} 
\hspace{-1.5em}
\tikz[baseline]{
    \draw [line width=0.05em]
        (0, 0.5em) -- 
        (0.45em, 0.5em) --
        (0.45em, 0.13333em) --
        (0.225em, -0.1em) --
        (0, 0.13333em) --
        cycle
}
\hspace{-0.65em}}}\xspace$.
In particular, we either select a random action (``explore'') or select the current best action that is allowed by the shield (``exploit'').
The selected action is then executed in the environment MDP~$M$.
Its output is recorded in the transition database~$D$ and used along with the immediate reward~$r$ to update the Q-table.
Every~$u$ episodes, we update the model estimate and shield.
After exceeding the training budget of~$N$ episodes, we return the final versions of the policy, the shield, and the model estimate.

\section{Experimental Evaluation}\label{sec:evaluation}

In this section, we evaluate our proposed adaptive probabilistic shielding approach from different angles.
We aim to answer the following research questions:

\begin{questionenum}[leftmargin=3.5em]
    \item \label{sec:safetyandreward} Can our approach learn a safe and optimal policy?
    \item \label{sec:exploration} What is the effect on the environment exploration?
    \item \label{sec:modeldistance} Does the model estimate improve over time?
    \item \label{sec:kapparate} Do the model estimates become sufficiently precise?
    \item \label{sec:estimators} What is the impact of the specific model estimator?
    \item \label{sec:unshieldedexploration} Is unshielded exploration beneficial?
    \item \label{sec:updatefrequency} How often should the shield be updated?
    \item \label{sec:horizon} What is the impact of the shield lookahead ($h$)?
\end{questionenum}

\subsection{Implementation and Baseline Methods}

The implementation is available online \cite{brorholt_2026_21874278}. We train the agent using standard Q-learning~\cite{DBLP:journals/ml/WatkinsD92} with the hyperparameters $\alpha=0.1$ (learning rate), $\gamma=0.9$ (discount factor), and $\varepsilon = 0.05$ (exploration probability).
We use reward shaping to penalize safety violations, with the penalty varying by environment.
Instead, we separately record the number of episodes that were unsafe.

By default, we use the LUI estimator $E_{\textrm{LUI}}$ with prior strengths $[\underline{n}_i,\overline{n}_i] = [5, 10]$ and a (``pessimistic'') robust shield attitude with parameters $\theta=0.05$, $\kappa=0.01$, and $h = 100$, and update the estimate and shield every $u = 1000$ episodes.
We underline these defaults in the following figures and tables.

We compare to two baselines.
The first baseline is a standard unshielded RL agent trained with reward shaping; this baseline is expected to perform poorly in terms of safety due to the lack of a shield.
The second baseline is a shielded RL agent that uses a probabilistic shield computed with the same method but given the ground-truth MDP (which our method cannot access); this ``oracle'' baseline acts as a benchmark and is expected to outperform all other methods.

\subsection{Description of Environments}

In total, we consider five different environments with mixed safety and optimization objectives.
The first environment is described in previous literature while the remaining were developed as additional benchmarks for our problem setting.

The \textbf{aircraft} environment~\cite{DBLP:conf/nips/SuilenS0022,Kochenderfer15} ($|S|=1665$) represents a collision avoidance system of an aircraft that must keep a minimum vertical distance to another plane that passes horizontally.
The other plane changes altitude at random, and the agent's aircraft may fail to follow the instructions with a small probability.

The \textbf{antlion} environment ($|S|=400$) requires the agent (an ant) to circumnavigate a stationary predator in order to reach a goal on the other side.
Instead of moving in the intended direction, the agent may slip toward the predator, with the probability increasing with proximity to it.
Reaching the goal yields a reward, while taking a step has a cost that decreases with proximity to the goal.

The \textbf{sinkholes} environment ($|S|=400$) features multiple goals with varying rewards and multiple holes that must be avoided.
If the agent falls into a hole, it either escapes with a small probability or returns to the starting state.
As the agent moves, it may slip in a random direction instead, with the probability varying by the state.
The cost of moving decreases with proximity to the goal.

The \textbf{crossroads} environment ($|S|=202$) illustrates deferred risk.
In the initial state, the agent must choose between two roads, which are then followed for $100$ steps.
One route is safe, while the other route is more rewarding but has a risk of slipping into an unsafe state at each step.

The \textbf{gravity} environment ($|S|=2000$) rewards the agent for visiting a sequence of checkpoints near a gravity well without crashing into the latter.
Every step has a small cost, and there is a probability that the agent is dragged towards the well, which increases with proximity.
Later checkpoints are riskier to visit, and the agent can end the episode early by going to one of two exit points.

\subsection{Experimental Results and Discussion of Research Questions}

Our experiments consists of $21$ hyperparameter configurations (e.g., the choice of the model estimator), and we repeated each run $100$ times.

\paragraph{\ref{sec:safetyandreward}: Can our approach learn a safe and optimal policy?}

\begin{figure}[tbp]
  \centering
  \hspace*{6mm} I \hspace*{25mm} II \hspace*{26.5mm} III \hspace*{24.5mm} IV \\[2mm]

  \includegraphics[width=0.24\linewidth]{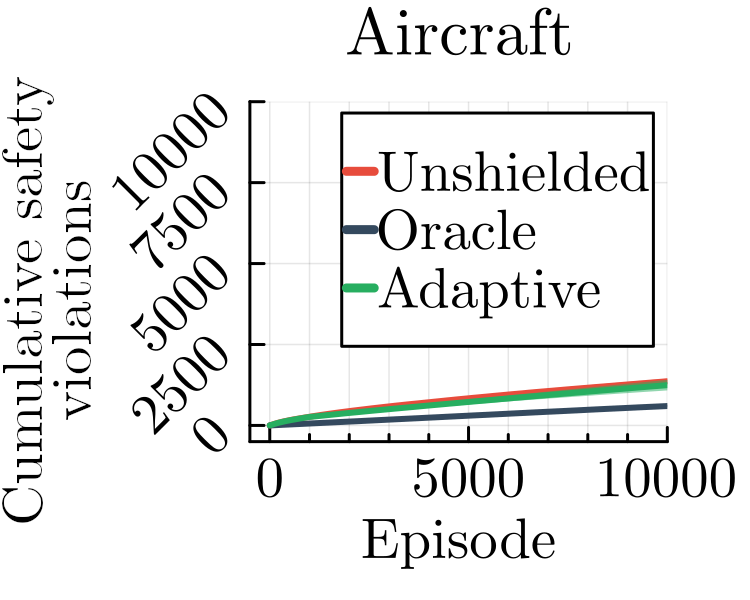}
  \includegraphics[width=0.24\linewidth]{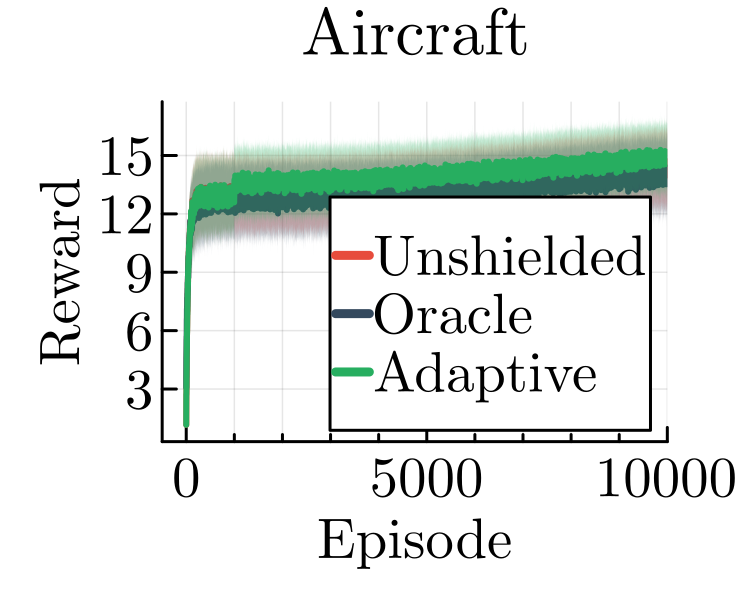}
  \includegraphics[width=0.24\linewidth]{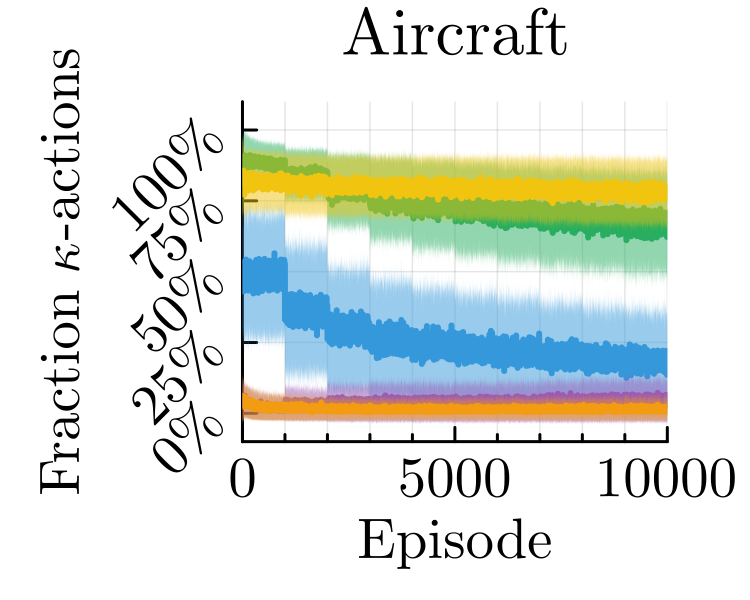}
  \includegraphics[width=0.24\linewidth]{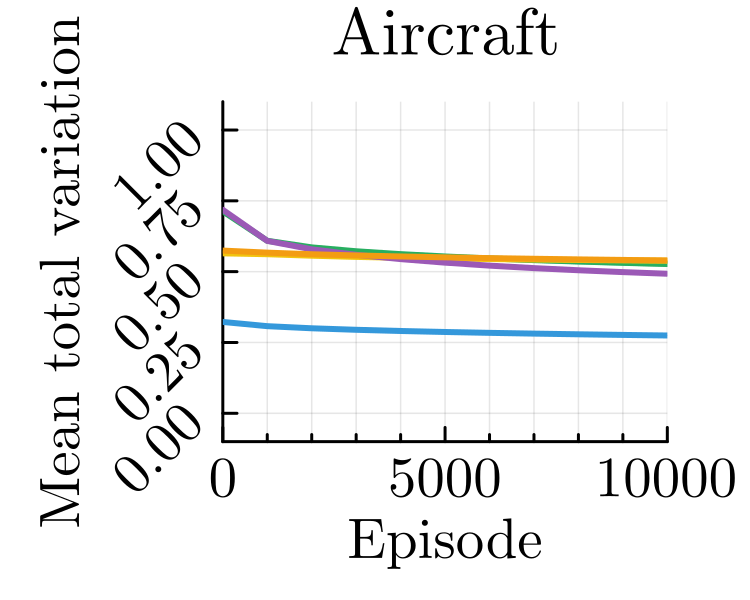}

  \includegraphics[width=0.24\linewidth]{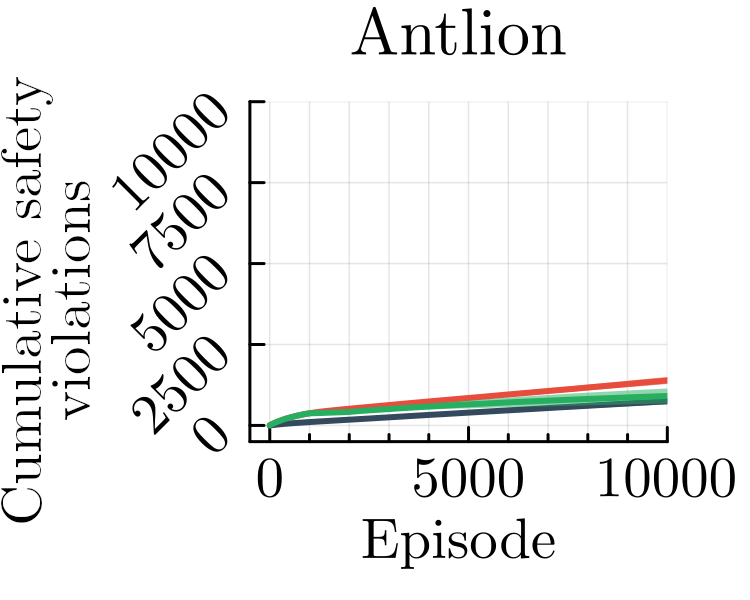}
  \includegraphics[width=0.24\linewidth]{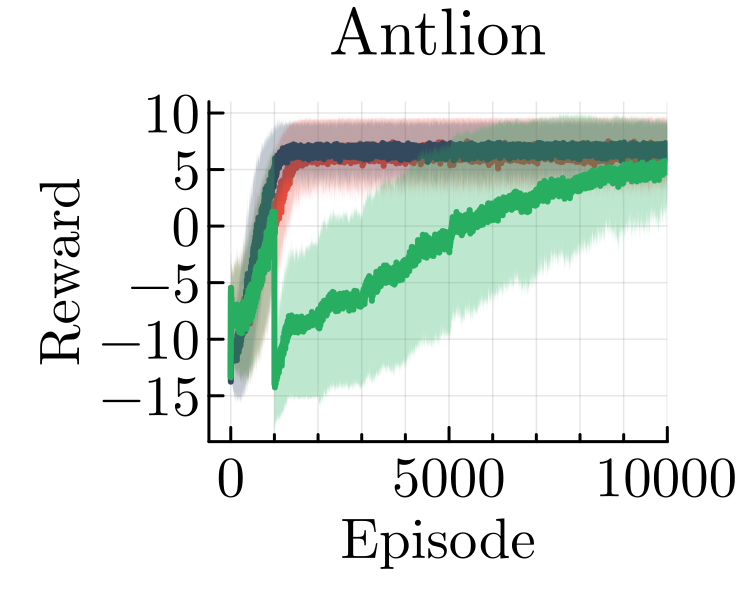}
  \includegraphics[width=0.24\linewidth]{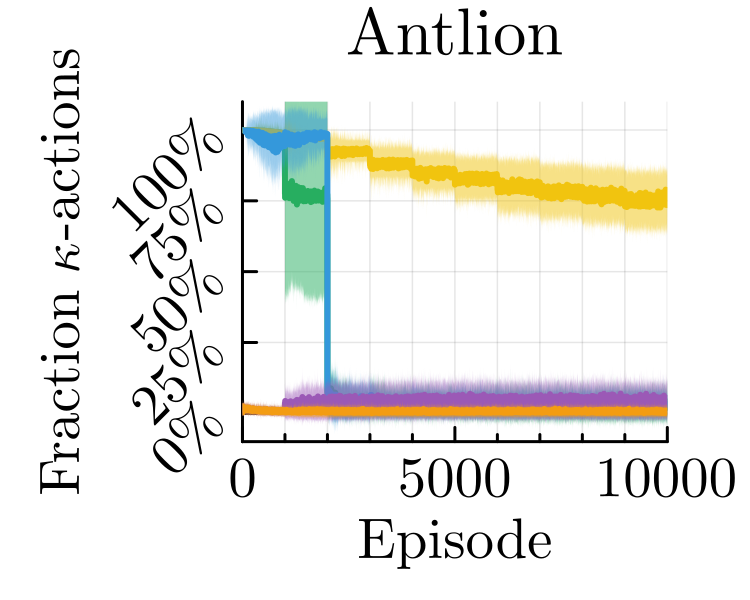}
  \includegraphics[width=0.24\linewidth]{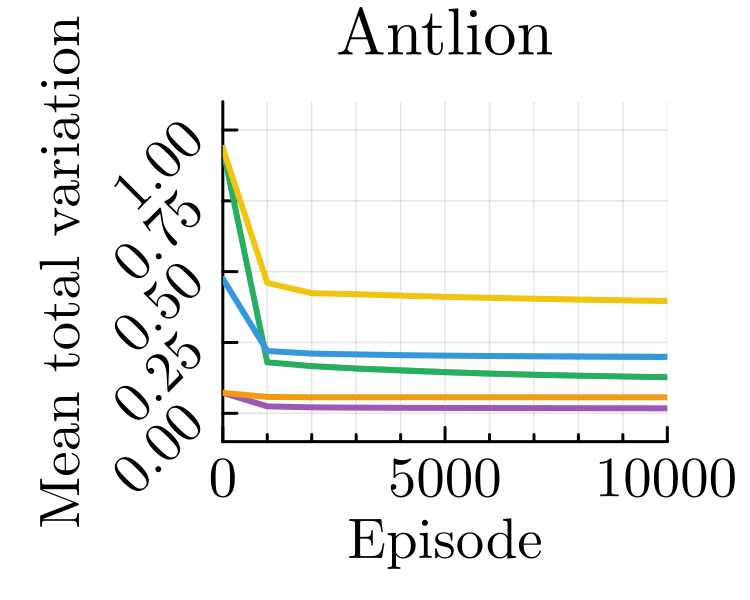}

  \includegraphics[width=0.24\linewidth]{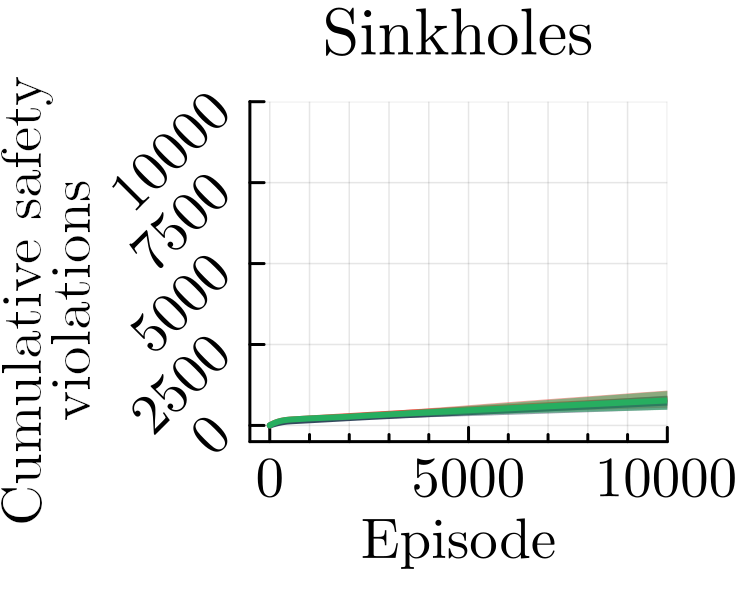}
  \includegraphics[width=0.24\linewidth]{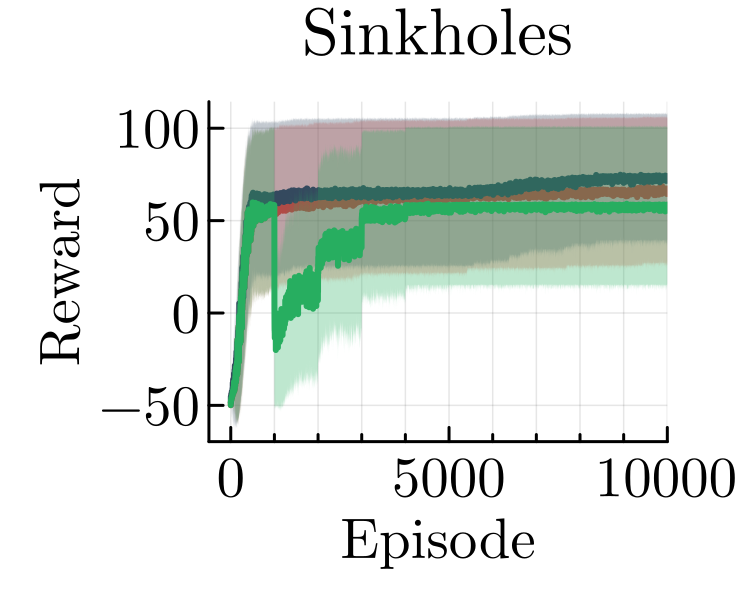}
  \includegraphics[width=0.24\linewidth]{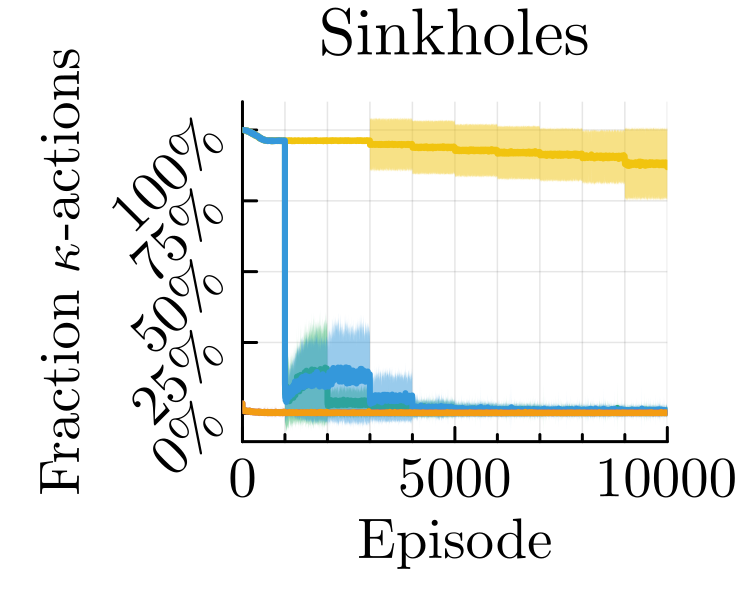}
  \includegraphics[width=0.24\linewidth]{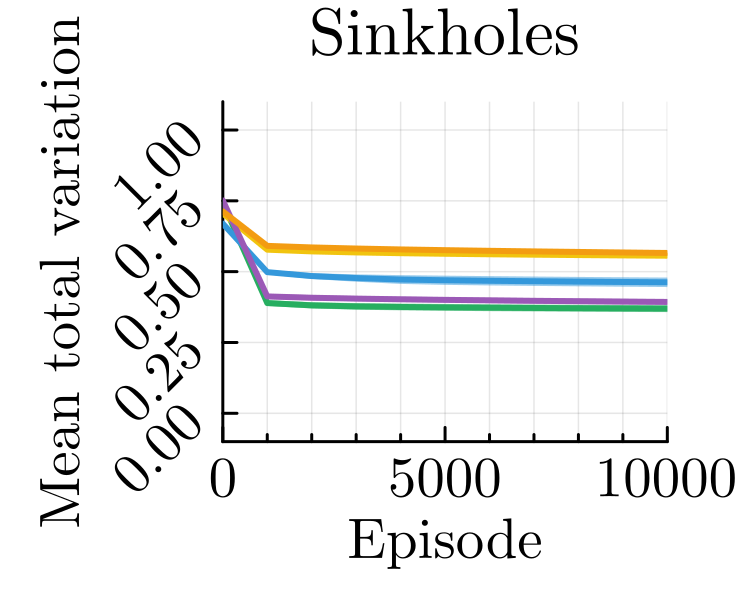}

  \includegraphics[width=0.24\linewidth]{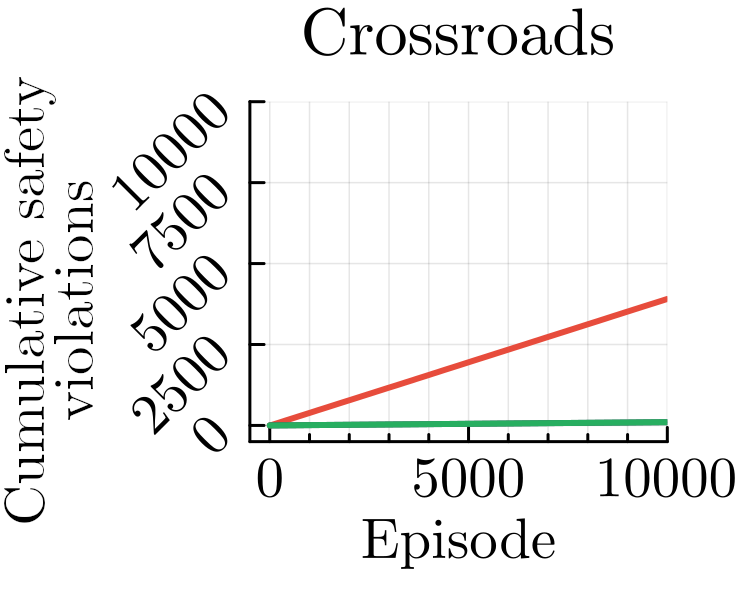}
  \includegraphics[width=0.24\linewidth]{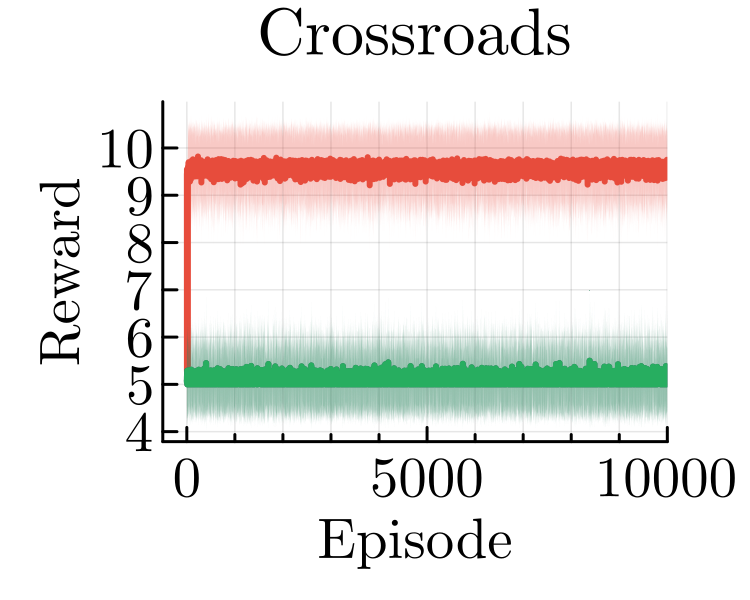}
  \includegraphics[width=0.24\linewidth]{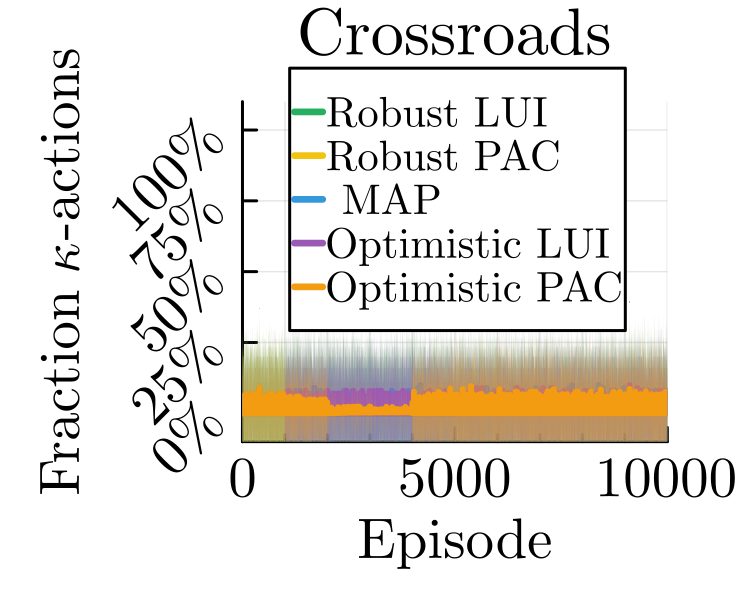}
  \includegraphics[width=0.24\linewidth]{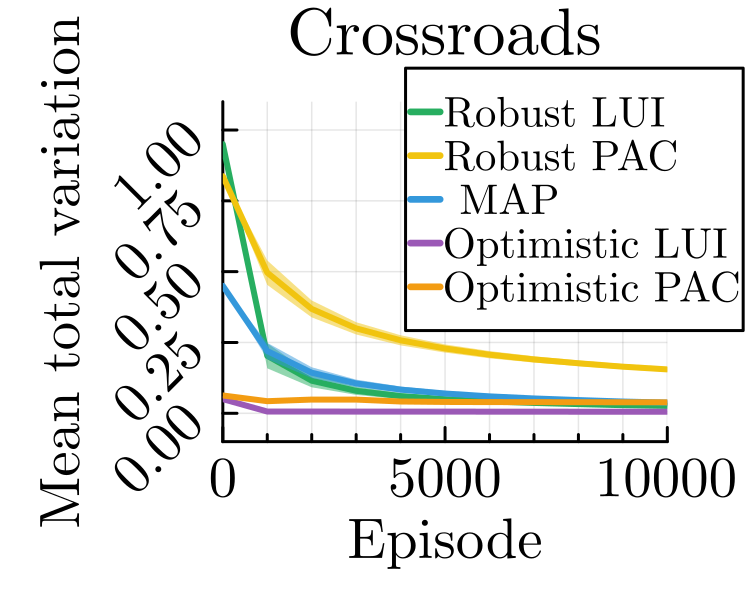}

  \includegraphics[width=0.24\linewidth]{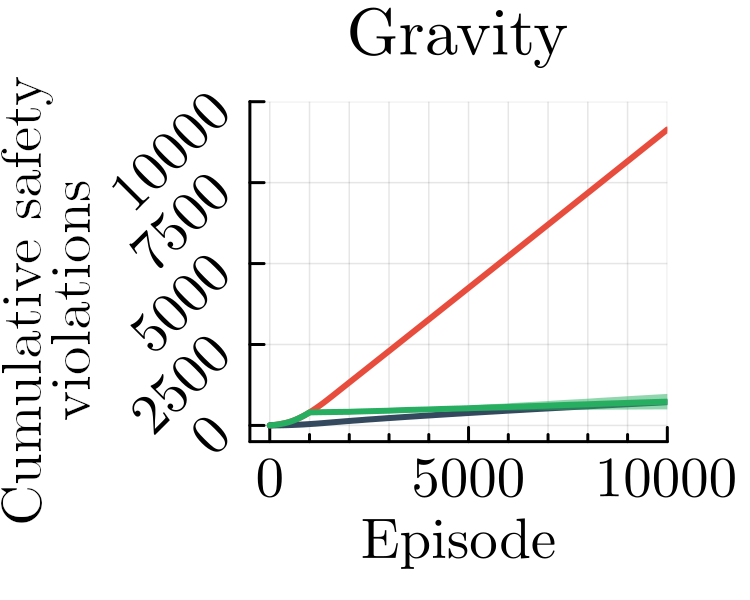}
  \includegraphics[width=0.24\linewidth]{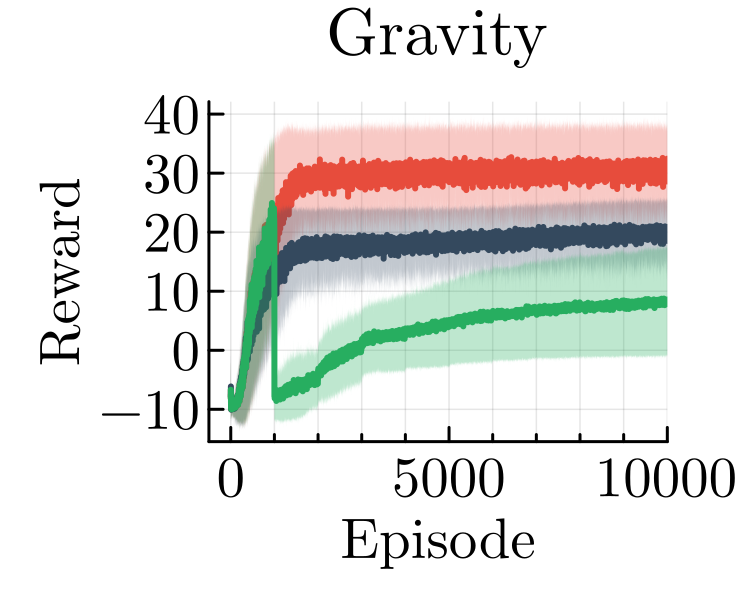}
  \includegraphics[width=0.24\linewidth]{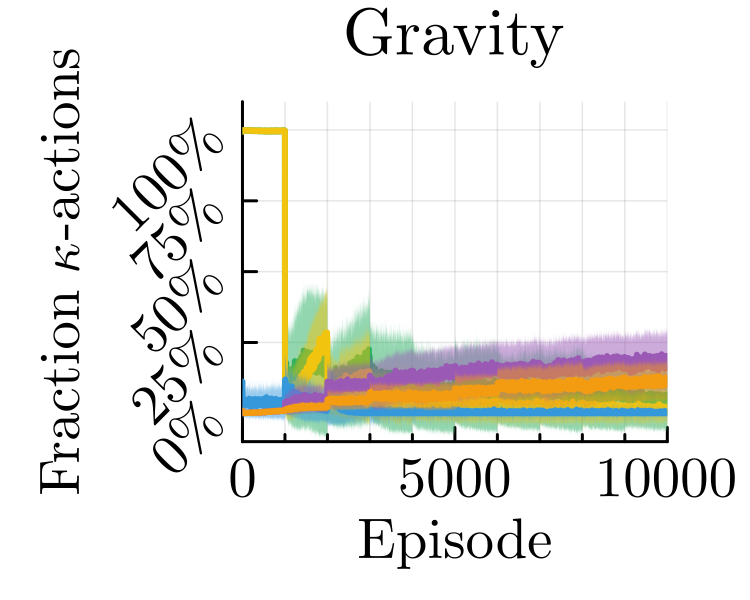}
  \includegraphics[width=0.24\linewidth]{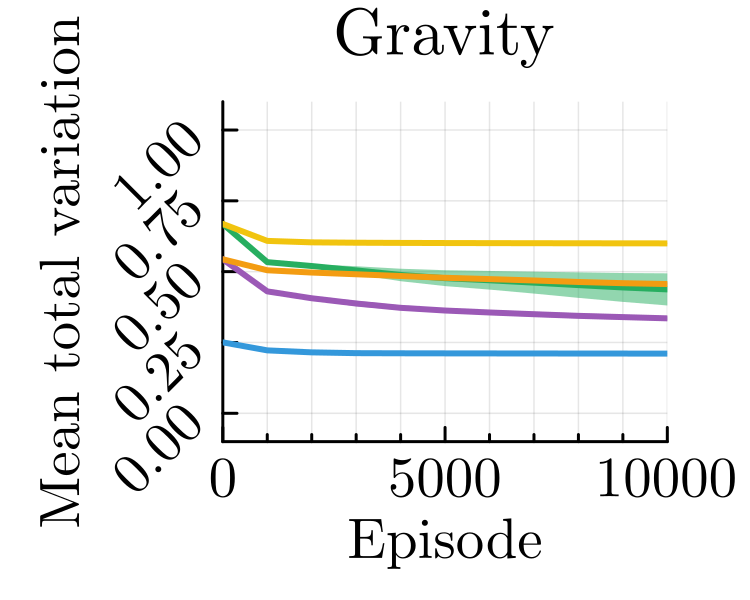}

  \caption{
    Mean outcome of 100 repetitions for different configurations.
    We plot the standard deviation as a ribbon around the lines.
    Vertical grid lines mark updates of the adaptive shield.
    Column~I: Cumulative safety violations during training.
    Column~II: Reward during training.
    Column~III: Per-episode rate of using the fallback~$\ensuremath{\text{\protect \input{tikz/Shield.tex}}}\xspace_\kappa$.
    Column~IV: Average total variation between true model and model estimate during training.
    }
    \label{fig:thefigure}
\end{figure}

\begin{figure}[tbp]
    \center
    \includegraphics[width=0.17\linewidth]{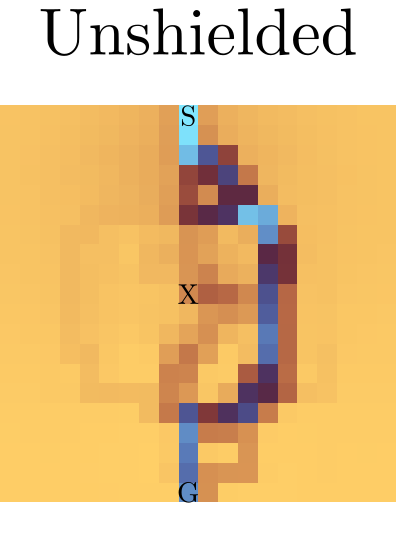}
    \includegraphics[width=0.17\linewidth]{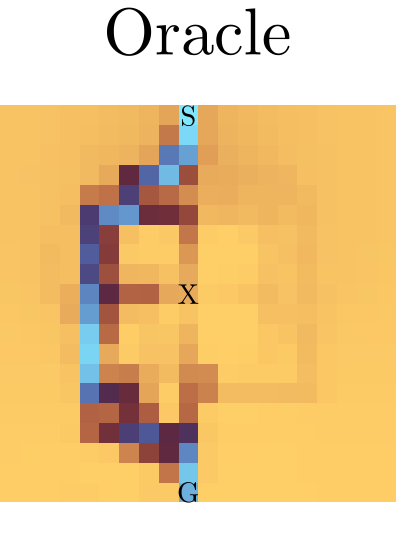}
    \includegraphics[width=0.17\linewidth]{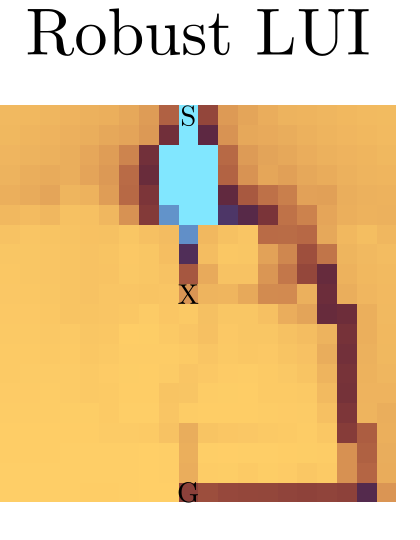}
    \includegraphics[width=0.17\linewidth]{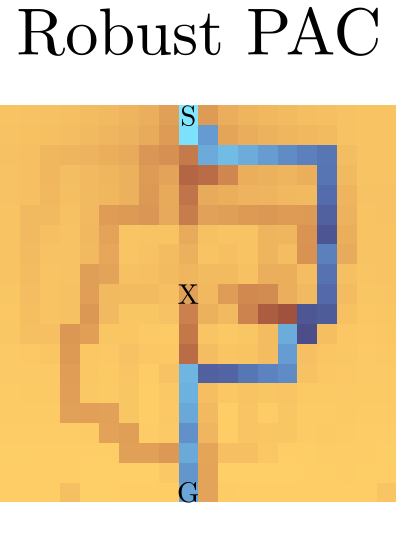}
    \includegraphics[width=0.17\linewidth]{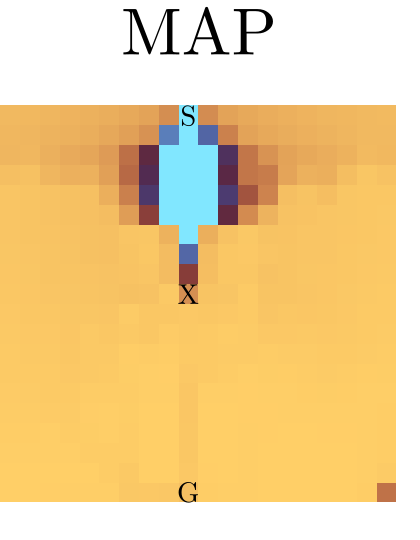}
    \includegraphics[width=0.08\linewidth,height=23mm,keepaspectratio]{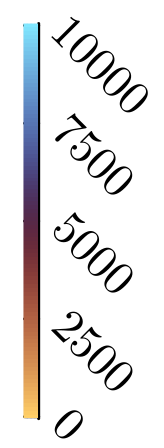}

    \caption{Heat map showing how frequently a state is visited in the antlion environment.
    The initial state, predator, and goal are at the bottom, center, and top, respectively.}
    \label{fig:heatmap}
\end{figure}

We are interested in both safety (\cref{fig:thefigure} column~I) and performance (\cref{fig:thefigure} column~II) over time.
(Note that the penalty stemming from the reward shaping is not included in these plots.)
For most environments, the adaptive shield leads to about the same number of safety violations as the oracle shield.
The unshielded baseline is generally less safe, especially in the crossroads and gravity environments, despite a strong reward penalty.
Still, the
negative outcomes were overshadowed by the more frequent positive rewards in the Q-learning algorithm.
We also note that the oracle baseline can generally explore more freely than the adaptive method because of its less conservative shield.
Yet, in the aircraft environment, the adaptive method actually achieves a slightly more rewarding policy,
profiting from slightly riskier behavior.
This may seem counterintuitive given the robust attitude of the shield, which is generally more conservative than the oracle baseline.
The reason we still see this behavior is that, during the exploration, profitable states are visited more often, making these seem safer as compared to less explored states with wider interval estimates.
Indeed, column~III reveals that the shield mostly falls back to the $\ensuremath{\text{\protect 
{\color{white} $\nabla$} 
\hspace{-1.5em}
\tikz[baseline]{
    \draw [line width=0.05em]
        (0, 0.5em) -- 
        (0.45em, 0.5em) --
        (0.45em, 0.13333em) --
        (0.225em, -0.1em) --
        (0, 0.13333em) --
        cycle
}
\hspace{-0.65em}}}\xspace_\kappa$ variant in this case.

Updates to the adaptive shield can be seen to temporarily affect performance negatively in the antlion, sinkholes, and gravity environments. 
At the $1000$ episode mark, the shield is updated for the first time, leading to a drop in the reward performance.
This drop is to be expected: as seen in column~III, the initial shield typically uses the $\ensuremath{\text{\protect 
{\color{white} $\nabla$} 
\hspace{-1.5em}
\tikz[baseline]{
    \draw [line width=0.05em]
        (0, 0.5em) -- 
        (0.45em, 0.5em) --
        (0.45em, 0.13333em) --
        (0.225em, -0.1em) --
        (0, 0.13333em) --
        cycle
}
\hspace{-0.65em}}}\xspace_\kappa$ variant because every action seems unsafe.
The first model update is most impactful and hence typically leads to a very different shield, and hence the agent effectively experiences a different environment.
Over time, the reward performance recovers as the policy adapts to the new shield.

Overall, the shields have a significant positive impact on safety and explore the environment more safely.
This safety may come at a cost when risky behavior is profitable, but this is a desirable trade-off in many applications.

\paragraph{\ref{sec:exploration}: What is the effect on the environment exploration?}

To assess how adaptive shields affect exploration, \cref{fig:heatmap} visualizes the number of times a state has been visited in the antlion environment during one algorithm execution over $10{,}000$ episodes, comparing the unshielded and oracle baselines as well as adaptive shields with the estimators \emph{robust LUI} (default), \emph{robust PAC}, and \emph{MAP}.

The LUI and PAC interval estimators both find the goal by following some narrow paths during learning. 
Meanwhile, the MAP estimator almost never leaves the area around the initial states.
This is consistent with the low average performance of the MAP estimator in \cref{tab:estimators} and further investigated below in \ref{sec:estimators}.
The unshielded agent passes close to the antlion but takes a more circuitous route, similar to what the oracle shield allows.
Both appear to explore more freely around the paths they take, rather than staying on a narrow route permitted by the adaptive shield.
This highlights the need for exploring states outside of what the shield allows, as also investigated in \ref{sec:unshieldedexploration} below.

\paragraph{\ref{sec:modeldistance}: Does the model estimate improve over time?}

We plot mean \emph{total variation}~(TV), i.e., $\nicefrac{1}{|S||A|}\sum_{s,a} \nicefrac{1}{2} \sum_{s' \in S} |{T^{\dagger}}(s,a,s') - T^*(s,a,s')|$, between the true MDP's transition probabilities $T^*$ and the probabilities $T^{\dagger}$ returned by PMC used for the shield (e.g., for robust/optimistic shields, as found from \cref{eq:imdpshield}).
We omit transitions with probability 1, since their probability is known.

As the agent explores the environment, we obtain a more precise estimate of the model.
Exploration also comes with risk; thus, it is not desirable to obtain a perfect estimate of all transitions -- in particular not those transitions that are rarely visited by the policy.
The mean TV after each model update is shown in column~IV of \cref{fig:thefigure}.
The most significant change occurs in the first update after $1000$ episodes,
indicating that this is sufficient to collect data for key transitions, which will be further explored during the rest of the training.
While the estimates for these transitions keep improving, this has little impact on the (global) metric.

\paragraph{\ref{sec:kapparate}: Do the model estimates become sufficiently precise?}

We examine how often the shield allows an action because it satisfies the $\theta$ threshold ($\ensuremath{\text{\protect 
{\color{white} $\nabla$} 
\hspace{-1.5em}
\tikz[baseline]{
    \draw [line width=0.05em]
        (0, 0.5em) -- 
        (0.45em, 0.5em) --
        (0.45em, 0.13333em) --
        (0.225em, -0.1em) --
        (0, 0.13333em) --
        cycle
}
\hspace{-0.65em}}}\xspace_\theta$) respectively how often it has to use the fallback ($\ensuremath{\text{\protect 
{\color{white} $\nabla$} 
\hspace{-1.5em}
\tikz[baseline]{
    \draw [line width=0.05em]
        (0, 0.5em) -- 
        (0.45em, 0.5em) --
        (0.45em, 0.13333em) --
        (0.225em, -0.1em) --
        (0, 0.13333em) --
        cycle
}
\hspace{-0.65em}}}\xspace_\kappa$) (cf.\ \cref{eq:shield}).
Column~III of \cref{fig:thefigure} shows the fallback frequency.
Until the first model update ($1000$ episodes), the robust and MAP estimates primarily use $\ensuremath{\text{\protect 
{\color{white} $\nabla$} 
\hspace{-1.5em}
\tikz[baseline]{
    \draw [line width=0.05em]
        (0, 0.5em) -- 
        (0.45em, 0.5em) --
        (0.45em, 0.13333em) --
        (0.225em, -0.1em) --
        (0, 0.13333em) --
        cycle
}
\hspace{-0.65em}}}\xspace_\kappa$, whereas the optimistic estimates, which consider almost all actions safe from the beginning, primarily use $\ensuremath{\text{\protect 
{\color{white} $\nabla$} 
\hspace{-1.5em}
\tikz[baseline]{
    \draw [line width=0.05em]
        (0, 0.5em) -- 
        (0.45em, 0.5em) --
        (0.45em, 0.13333em) --
        (0.225em, -0.1em) --
        (0, 0.13333em) --
        cycle
}
\hspace{-0.65em}}}\xspace_\theta$.
The slowest improvement of the estimate is observed in the aircraft environment.
This is because, unlike the static obstacles in the other environments, learning the behavior of the randomly moving opponent requires more data.
In all other environments, $\ensuremath{\text{\protect 
{\color{white} $\nabla$} 
\hspace{-1.5em}
\tikz[baseline]{
    \draw [line width=0.05em]
        (0, 0.5em) -- 
        (0.45em, 0.5em) --
        (0.45em, 0.13333em) --
        (0.225em, -0.1em) --
        (0, 0.13333em) --
        cycle
}
\hspace{-0.65em}}}\xspace_\theta$ is used most of the time after one or two model updates for all but the robust PAC estimator, which sometimes fails to obtain a sufficiently precise estimate (due to its higher data requirements).

\paragraph{\ref{sec:estimators}: What is the impact of the specific model estimator?}

\begin{table}[tb]
    \caption{Comparison for different model estimators.
    The numbers in each cell respectively denote the reward in the final evaluation (left) and the probability that the final policy produces an unsafe episode (right). The numbers are the mean outcomes of $100$ repetitions and bold entries mark the best result for each column.}
    \label{tab:estimators}
    \centering
    \begin{tabular}{c @{\tsp} c @{\tsp} c @{\tsp} c @{\tsp} c @{\tsp} c}
        \toprule
        Estimator & Aircraft & Antlion & Sinkholes & Crossroads & Gravity \\
        \midrule
        \underline{Robust LUI} & \makecell{14.89 \csp 8.3\%} & \makecell{5.27 \csp 4.5\%} & \makecell{57.05 \csp 3.6\%} & \makecell{5.12 \csp \textbf{0.0\%}} & \makecell{8.27 \csp 3.9\%} \\
        Robust PAC & \makecell{\textbf{15.54} \csp 8.2\%} & \makecell{6.33 \csp \textbf{2.8\%}} & \makecell{46.59 \csp 3.8\%} & \makecell{5.12 \csp \textbf{0.0\%}} & \makecell{$-$2.56 \csp \textbf{0.0\%}} \\
        MAP & \makecell{13.24 \csp \textbf{4.0\%}} & \makecell{$-$8.99 \csp 8.0\%} & \makecell{49.16 \csp \textbf{3.0\%}} & \makecell{5.12 \csp \textbf{0.0\%}} & \makecell{$-$2.35 \csp \textbf{0.0\%}} \\
        Optimistic LUI & \makecell{14.17 \csp 5.2\%} & \makecell{6.66 \csp 6.0\%} & \makecell{67.78 \csp 4.1\%} & \makecell{5.11 \csp \textbf{0.0\%}} & \makecell{21.40 \csp 19.6\%} \\
        Optimistic PAC & \makecell{14.35 \csp 6.8\%} & \makecell{6.36 \csp 10.0\%} & \makecell{65.35 \csp 3.8\%} & \makecell{5.12 \csp \textbf{0.0\%}} & \makecell{23.51 \csp 47.8\%} \\
        \midrule
        Unshielded & \makecell{14.36 \csp 7.3\%} & \makecell{6.62 \csp 9.0\%} & \makecell{65.62 \csp 3.7\%} & \makecell{\textbf{9.57} \csp 40.1\%} & \makecell{\textbf{30.35} \csp 99.2\%} \\
        Oracle & \makecell{13.98 \csp 4.1\%} & \makecell{\textbf{6.78} \csp 5.6\%} & \makecell{\textbf{73.32} \csp 3.8\%} & \makecell{5.12 \csp \textbf{0.0\%}} & \makecell{19.86 \csp 4.5\%} \\
        \bottomrule
    \end{tabular}
\end{table}

\begin{table}[tbp]
    \caption{Comparison for the two exploration variants.
    Setup as in \cref{tab:estimators}.}
    \label{tab:exploration}
    \centering
    \begin{tabular}{c @{\tsp} c @{\tsp} c @{\tsp} c @{\tsp} c @{\tsp} c}
        \toprule
        Exploration & Aircraft & Antlion & Sinkholes & Crossroads & Gravity \\
        \midrule
        \underline{$\mathrm{Unif}_A$} & \makecell{14.89 \csp \textbf{8.3\%}} & \makecell{\textbf{5.27} \csp \textbf{4.5\%}} & \makecell{57.05 \csp \textbf{3.6\%}} & \makecell{\textbf{5.12} \csp \textbf{0.0\%}} & \makecell{\textbf{8.27} \csp 3.9\%} \\
        $\mathrm{Unif}_{\ensuremath{\text{\protect 
{\color{white} $\nabla$} 
\hspace{-1.5em}
\tikz[baseline]{
    \draw [line width=0.05em]
        (0, 0.5em) -- 
        (0.45em, 0.5em) --
        (0.45em, 0.13333em) --
        (0.225em, -0.1em) --
        (0, 0.13333em) --
        cycle
}
\hspace{-0.65em}}}\xspace(s)}$ & \makecell{\textbf{15.04} \csp 9.8\%} & \makecell{1.93 \csp 5.1\%} & \makecell{\textbf{57.29} \csp 3.7\%} & \makecell{5.00 \csp \textbf{0.0\%}} & \makecell{1.58 \csp \textbf{0.6\%}} \\
        \bottomrule
    \end{tabular}
\end{table}

We investigate how the different model estimators impact the results.
Specifically, we compare the iMDP estimators PAC and LUI, both with the robust and the optimistic attitudes, and the MDP estimator MAP.
We fix the priors of MAP to $w=10$ and of LUI to $[\underline{n}_i,\overline{n}_i] = [5, 10]$, and the parameters of PAC to $\delta=0.1$ and $\xi=10^{-8}$.
Varying priors as an additional dimension of experimental parameters is left for future research.

In \cref{tab:estimators}, we show the evaluation of the final policies.
Each cell shows the reward, evaluated empirically as the mean over $1000$ episodes, and the relative safety of this policy, computed analytically using the PRISM model checker~\cite{PRISM}.\footnote{%
We did not use PRISM to compute the reward because it does not support a mix of positive and negative rewards.
}

Counterintuitively, robust estimators do not always lead to safer policies, as seen in the aircraft
environment.
Since the agent initially finds an imperfect route, which is considered relatively safe compared to less-explored states, the shield later forces the agent to stay on it, even if an unexplored yet safer alternative may exist.
Instead, the optimistic shield specifications allow the agent to explore more states unless prior experience indicates that doing so is unsafe.
The sinkholes and gravity environments show the biggest variation in the results.
This is because the goals (checkpoints) in the sinkholes (gravity) environment are far apart, and hence finding a good route strongly depends on the exploration.

\paragraph{\ref{sec:unshieldedexploration}: Is unshielded exploration beneficial?}

In Line~\ref{ln:explore} of \cref{alg:alg}, the $\varepsilon$-greedy exploration chooses a random action from the set of all actions~$\mathrm{Unif}_A$, thus disregarding the shield.
In \cref{tab:exploration}, we compare this strategy to the alternative that only admissible actions are allowed in state~$s$ with $\mathrm{Unif}_{\ensuremath{\text{\protect 
{\color{white} $\nabla$} 
\hspace{-1.5em}
\tikz[baseline]{
    \draw [line width=0.05em]
        (0, 0.5em) -- 
        (0.45em, 0.5em) --
        (0.45em, 0.13333em) --
        (0.225em, -0.1em) --
        (0, 0.13333em) --
        cycle
}
\hspace{-0.65em}}}\xspace(s)}$.
Generally, with the latter variant, the agent eventually stops exploring new actions once the shield has found at least one admissible route.
For some environments, this change has no significant effect on the reward because the route that was identified first was sufficiently good, while in the antlion and gravity environments, the restricted variant prevents the agent from uncovering more promising routes.

\paragraph{\ref{sec:updatefrequency}: How often should the shield be updated?}

\begin{table}[tbp]
    \caption{Comparison for different update delays~$u$.
    Setup as in \cref{tab:estimators}.}
    \label{tab:frequency}
    \centering
    \begin{tabular}{c @{\tsp} c @{\tsp} c @{\tsp} c @{\tsp} c @{\tsp} c}
        \toprule
        $u$ & Aircraft & Antlion & Sinkholes & Crossroads & Gravity \\
        \midrule
        \phantom{0\,}250 & \makecell{14.94 \csp 12.9\%} & \makecell{$-$10.74 \csp 9.7\%} & \makecell{78.71 \csp 5.4\%} & \makecell{\textbf{5.12} \csp \textbf{0.0\%}} & \makecell{$-$2.32 \csp \textbf{0.0\%}} \\
        \phantom{0\,}500 & \makecell{\textbf{14.99} \csp 10.5\%} & \makecell{$-$10.46 \csp 11.0\%} & \makecell{\textbf{84.78} \csp 4.1\%} & \makecell{\textbf{5.12} \csp \textbf{0.0\%}} & \makecell{2.95 \csp 1.9\%} \\
        \underline{1\,000} & \makecell{14.89 \csp 8.3\%} & \makecell{5.27 \csp 4.5\%} & \makecell{57.05 \csp \textbf{3.6\%}} & \makecell{\textbf{5.12} \csp \textbf{0.0\%}} & \makecell{8.27 \csp 3.9\%} \\
        1\,500 & \makecell{14.81 \csp 7.8\%} & \makecell{\textbf{6.17} \csp \textbf{3.7\%}} & \makecell{54.40 \csp \textbf{3.6\%}} & \makecell{\textbf{5.12} \csp \textbf{0.0\%}} & \makecell{11.02 \csp 3.5\%} \\
        2\,000 & \makecell{14.72 \csp \textbf{7.0\%}} & \makecell{6.06 \csp 4.0\%} & \makecell{57.93 \csp \textbf{3.6\%}} & \makecell{\textbf{5.12} \csp \textbf{0.0\%}} & \makecell{\textbf{12.92} \csp 8.2\%} \\
        \bottomrule
    \end{tabular}
\end{table}

\begin{table}[tbp]
    \caption{Comparison for varying shield horizon ($h$ in $\varphi|h$).
    Setup as in \cref{tab:estimators}.}
    \label{tab:horizon}
    \centering
    \begin{tabular}{c @{\tsp} c @{\tsp} c @{\tsp} c @{\tsp} c @{\tsp} c}
        \toprule
        $h$ & Aircraft & Antlion & Sinkholes & Crossroads & Gravity \\
        \midrule
        \phantom{00}6 & \makecell{14.50 \csp \textbf{8.2\%}} & \makecell{\textbf{6.42} \csp \textbf{3.1\%}} & \makecell{48.50 \csp \textbf{3.3\%}} & \makecell{\textbf{9.57} \csp 40.1\%} & \makecell{\textbf{24.61} \csp 47.1\%} \\
        \phantom{0}12 & \makecell{14.78 \csp \textbf{8.2\%}} & \makecell{5.90 \csp 3.8\%} & \makecell{54.81 \csp 3.4\%} & \makecell{\textbf{9.57} \csp 40.1\%} & \makecell{23.16 \csp 33.3\%} \\
        \phantom{0}25 & \makecell{\textbf{14.89} \csp 8.3\%} & \makecell{5.71 \csp 4.2\%} & \makecell{56.31 \csp 3.7\%} & \makecell{\textbf{9.57} \csp 40.1\%} & \makecell{22.16 \csp 21.5\%} \\
        \phantom{0}50 & \makecell{\textbf{14.89} \csp 8.3\%} & \makecell{4.67 \csp 4.9\%} & \makecell{57.12 \csp 3.4\%} & \makecell{\textbf{9.57} \csp 40.1\%} & \makecell{19.36 \csp 13.2\%} \\
        \phantom{0}75 & \makecell{\textbf{14.89} \csp 8.3\%} & \makecell{4.39 \csp 4.4\%} & \makecell{56.35 \csp 3.7\%} & \makecell{\textbf{9.57} \csp 40.1\%} & \makecell{13.55 \csp 6.9\%} \\
        \underline{100} & \makecell{\textbf{14.89} \csp 8.3\%} & \makecell{5.27 \csp 4.4\%} & \makecell{57.05 \csp 3.7\%} & \makecell{5.12 \csp \textbf{0.0\%}} & \makecell{8.27 \csp 4.0\%} \\
        125 & \makecell{\textbf{14.89} \csp 8.3\%} & \makecell{5.37 \csp 4.0\%} & \makecell{57.09 \csp 3.7\%} & \makecell{5.12 \csp \textbf{0.0\%}} & \makecell{5.21 \csp 2.7\%} \\
        150 & \makecell{\textbf{14.89} \csp 8.3\%} & \makecell{5.14 \csp 3.6\%} & \makecell{57.07 \csp 3.9\%} & \makecell{5.12 \csp \textbf{0.0\%}} & \makecell{3.81 \csp 2.5\%} \\
        175 & \makecell{\textbf{14.89} \csp 8.3\%} & \makecell{5.42 \csp 3.6\%} & \makecell{\textbf{57.19} \csp 3.7\%} & \makecell{5.12 \csp \textbf{0.0\%}} & \makecell{2.90 \csp 2.0\%} \\
        200 & \makecell{\textbf{14.89} \csp 8.3\%} & \makecell{5.33 \csp \textbf{3.1\%}} & \makecell{57.18 \csp 3.6\%} & \makecell{5.12 \csp \textbf{0.0\%}} & \makecell{1.24 \csp \textbf{1.5\%}} \\
        \bottomrule
    \end{tabular}
\end{table}

The model estimate and subsequent shield update are the most expensive operations in our approach.
We examine the effect of the update delay~$u$.
\cref{tab:frequency} shows that the choice of~$u$ can be impactful, but no single choice is more preferable across the environments.

\paragraph{\ref{sec:horizon}: What is the impact of the shield lookahead ($h$)?}

We examine the effect of varying the lookahead horizon~$h$ of the shield.
\cref{tab:horizon} shows the results.
Aircraft episodes end after $20$ steps, making longer horizons redundant.
The gravity environment is less safe at lower lookahead.
In the crossroads environment, for $h \le 75$, the agent prefers the more rewarding (but less safe) route.

\section{Conclusion and Future Work}

In this paper, we have proposed the paradigm of \emph{adaptive probabilistic shielding} for safe reinforcement learning.
We assume only access to a nondeterministic environment model (i.e., we do not know the transition probabilities) and consequently do not have access to a simulator.
In settings where safety violations are costly, safe exploration is a challenge.
We tackle this challenge with a practical, integrated procedure that simultaneously explores the environment, updates a model estimate, maintains a probabilistic shield, and learns a policy under that shield.
Our focus has been on the empirical evaluation, in which we investigated various research questions regarding the success and impact of our design choices.

While our shield implementation is based on a recent approach for interval MDPs~\cite{galesloot2026robustprobabilisticshieldingsafe}, the procedure can be extended to support other shields.
One direction is to vary the safety thresholds of the shields computed during the course of the algorithm, which we have kept to a user-defined constant in our approach.
For instance, we imagine first using an optimistic shield that encourages exploration and then gradually raising the safety threshold to obtain a safer shield in the end.
As another direction,
one can
incorporate more elements from model-based RL algorithms.
For instance, we may replace the uniformly random exploration step with a biased choice toward under-explored states and actions.

\begin{credits}
\subsubsection{\ackname} This research was partly supported by the European Research Council (ERC) Starting Grant 101077178 (DEUCE), the Villum Investigator Grant S4OS under reference number 37819, and the Independent Research Fund Denmark under reference number 10.46540/3120-00041B.
 
\subsubsection{\discintname}
The authors have no competing interests to declare that are relevant to the content of this article. \end{credits}

%
%
\bibliographystyle{splncs04}
\bibliography{mybibliography}

\end{document}